\documentclass[nouppercase]{kaia}
\usepackage{amsmath}
\usepackage{amssymb}
\usepackage{float}
\usepackage[T1]{fontenc}
\usepackage{cite}
\makeatletter
\renewcommand\thesection{\Roman{section}}

\renewcommand\p@subsection{\thesection-}
\renewcommand\@seccntformat[1]{\csname the#1\endcsname.\quad}
\makeatother
\hypersetup{hidelinks,
  pdfauthor={Jonghyun Lee, Soojung Kang},
  pdftitle={Weight Averaging at Small Budgets Confounds Formula-Labeled Pretraining in Microstructure-Property Regression: A Pre-Specified Study with Post Hoc Rechecks}}

\title{Weight Averaging at Small Budgets Confounds Formula-Labeled Pretraining in Microstructure--Property Regression: A Pre-Specified Study with Post Hoc Rechecks}

\author{Jonghyun Lee}{a}
\author{Soojung Kang}{b}
\affiliation{Yonsei University, South Korea}{a}
\affiliation{Kakao Piccoma Corp., South Korea}{b}
\email{devjh@yonsei.ac.kr, kangsuroro2907@gmail.com}

\begin{document}

\maketitle
\begin{abstract}
Closed-form physics formulas can label synthetic microstructures at no cost, which makes supervised pretraining on such labels an alternative to self-supervised pretraining when labels are scarce. We test this idea with one pre-specified protocol on two public simulated benchmarks, two-phase conductivity and short-fiber stiffness: a ConvNeXt-Small backbone is pretrained on 150,000 synthetic images by regressing a formula label or by masked reconstruction and is fine-tuned on 100, 400, or 1,600 benchmark labels. Under the pre-specified protocol, five fine-tuning seeds from one formula checkpoint have mean residual correlations of 0.994 (conductivity) and 0.999 (stiffness), and on conductivity the formula-pretrained network compresses its predictions (slope 0.64) and has four times the 400-label error of ImageNet initialization. Post hoc analyses identify two interacting confounds in the conductivity results: the synthetic generator renders the low-conductivity phase with the opposite contrast polarity to the benchmark, and the pre-specified exponential moving average (EMA) of weights, with decay 0.995 and no bias correction, still holds 16\% of the initial weights after the 400-label schedule. With the EMA disabled, the compression disappears (slope 0.97) and, over three label subsets and five fine-tuning seeds, the mismatched checkpoint has $1.4$ times the error of ImageNet initialization (0.0119 versus 0.0084); with three pretraining runs per polarity, the matched run means (0.0086, 0.0111, 0.0118) separate below the mismatched means (0.0119, 0.0120, 0.0123) in aggregate but overlap them on two of three label subsets, so under a pre-specified rule the comparison is inconclusive: the direction is consistent (matched mean 0.0105 versus mismatched 0.0121) but three runs give too little power, and the benefit is heterogeneous across matched runs (spread 0.0032 against 0.0004 for the mismatched runs). On stiffness without the EMA, formula pretraining lowers the 100-label error by 18\% relative to ImageNet over five seeds, also with a re-initialized head, whereas the near-unit stiffness seed correlation appears for ImageNet seeds as well (0.992). Weight averages that start at the pretrained weights can thus dominate small-budget comparisons of pretrained initializations, and effects measured under them should be rechecked at the last weights.
\end{abstract}
\begin{keywords}
Formula-labeled pretraining, synthetic data, transfer learning, ensemble diversity, microstructure--property linkage
\end{keywords}

\section{Introduction}
Image-to-scalar regression from micrographs to a property is a central task of materials informatics, and its main bottleneck is the number of labeled images: a property label usually requires a destructive test or a finite-element method (FEM) simulation, so public datasets contain between a few hundred and a few thousand labeled representative volume elements (RVEs). Two families of remedies exist. Self-supervised pretraining learns a representation from unlabeled images and then fine-tunes on the few real labels \cite{wei_2024_11,he_2022_mae}. Simulation-based pretraining instead generates synthetic microstructures with known labels, at the cost of running a numerical solver for every image \cite{liu_2025_4,subramanian_2023_5}.

A third route avoids per-image simulation. When a microstructure is produced by a generator, every geometric quantity of that image (phase fraction, inclusion aspect ratio, orientation) is known exactly, and closed-form homogenization formulas map such quantities to a property \cite{tucker_1999_8,li_2004_3,watt_1976_7}. One can therefore label 150,000 synthetic images in seconds and pretrain a regressor on them. Two questions about this approach have not been examined under controlled conditions: does a formula label transfer better than a label-free pretext on public data, and what happens to an ensemble whose members all learned the same formula?

This paper addresses both questions with a single, fixed protocol applied to two public benchmarks: the effective conductivity $\kappa_{11}$ of two-phase microstructures \cite{lissner_2019,lissner_darus} and the stiffness component $C_{1111}$ of short-fiber composites \cite{wei_2024_11,wei_zenodo}. Every hyperparameter was fixed before any test label was seen and was not revised afterward. The protocol was fixed in an internal project file, not in a public registry, so we call it pre-specified; its timing cannot be verified independently. One fixed choice, the exponential moving average (EMA) of weights used for prediction, was found to dominate the small-budget comparisons, so we report the pre-specified numbers next to a post hoc recheck without it.

The answers to the two questions are qualified. For the first, once the weight average is removed the formula label helps only at the smallest stiffness budget, an $18\%$ encoder-level gain over five seeds at 100 labels, and on conductivity it does not beat a strong label-free control (a ViT masked autoencoder) once the EMA is removed (one seed); matching the generator polarity helps on only one of three pretraining runs, a heterogeneous and, at three runs per side, inconclusive effect. For the second, the near-perfect seed correlation is not specific to formula pretraining: it arises from the weight average and from error shared across the task, and ImageNet seeds collapse as well. Each item below states which part is pre-specified and which is post hoc.

\begin{enumerate}
\item \textbf{Seed dependence} (pre-specified, rechecked post hoc). Five fine-tuning seeds from one formula checkpoint have mean residual correlations $\bar\rho$ of $0.994$ and $0.999$ (effective member count $N_{\mathrm{eff}}\le1.01$), and averaging gains at most $0.012$ in root-mean-square error (RMSE). Without the EMA, $\bar\rho$ falls to $0.73$ on conductivity and stays at $0.998$ on stiffness, where five ImageNet seeds also reach $0.992$; the stiffness value is largely set by the task, not by the formula checkpoint.
\item \textbf{Transfer} (pre-specified, rechecked post hoc). Formula pretraining is the best initialization for stiffness at 100 labels (RMSE $5.49$ versus $13.28$ for ImageNet; without the EMA, over five seeds, $4.58$ versus $5.55$, and $4.60$ with its head re-initialized) and is within $1.5\%$ at 400 and 1,600 labels. On conductivity, ImageNet initialization beats both pre-specified pretexts at every budget with the EMA ($0.0133$ versus $0.0528$ at 400 labels) and is better than every mismatched pretraining run by run means without it ($0.0084$ versus $0.0119$--$0.0123$ over three runs, three subsets, and five seeds; by seed range run 0 is mixed across subsets).
\item \textbf{Polarity} (post hoc). The conductivity generator inverts the contrast polarity of the benchmark. With the EMA, matching it lowers the 400-label RMSE to $0.0170$--$0.0208$ over three pretraining runs. Without the EMA, with three pretraining runs per polarity, the matched run means ($0.0086$, $0.0111$, $0.0118$) separate below the mismatched means ($0.0119$, $0.0120$, $0.0123$) in aggregate but overlap on two of three subsets, so a pre-specified rule is inconclusive: most of the effect seen with the EMA is the retained wrong-sign readout, and what remains is a consistent but underpowered direction (matched mean $0.0105$ versus $0.0121$) whose size is heterogeneous, large on some matched runs and near zero on others.
\item \textbf{Weight averaging} (post hoc). The EMA retains $64\%$ and $16\%$ of the initial weights at 100 and 400 labels and inflates every small-budget error, most for initializations far from the target; the masked-reconstruction failure at 100 conductivity labels ($0.741$, $R^2=-27$) becomes $0.054$ without it.
\item \textbf{Baselines} (pre-specified ridge, post hoc calibration). A post hoc affine-calibrated formula ($0.033$) and a pre-specified frozen DINOv2 ridge model ($0.024$) outperform every EMA-averaged mismatched-polarity ConvNeXt at 400 conductivity labels but none of the networks without the EMA.
\end{enumerate}

\section{Related Work}
Synthetic microstructures have been used to pretrain or augment models for segmentation \cite{ozaki_2025_1} and for property regression through masked-image reconstruction with a vision transformer (ViT) masked autoencoder \cite{wei_2024_11}. Formula-labeled pretraining differs from both because the pretext target is a property, not the image, and from formula-driven supervised learning \cite{kataoka_2020}, whose labels identify the generating formula rather than a physical quantity. Both benchmarks here are simulated RVEs with solver labels; transfer to experimental micrographs \cite{ozaki_2025_1} and to three-dimensional volumes is outside our scope. Domain randomization \cite{tobin_2017} varies rendering parameters so that the target looks like one more variation; our mixed-polarity run applies it to one parameter. Learning a correction to a physics baseline \cite{ramakrishnan_2015} is the natural use of an accurate formula, and our affine baseline is its simplest form. Shortcut learning, in which a network exploits a cue that is predictive in the training domain but absent or inverted in the target domain, is the general phenomenon underlying the confound we report \cite{geirhos_2020}.

Ensemble error has a precise decomposition into bias, variance, and diversity terms \cite{wood_2023_2}, and deep ensembles are known to lose diversity when members converge to similar functions \cite{deng_2022_9,vieira_2026_6}. Ensembles of models fine-tuned from one checkpoint tend to remain in one loss basin and to gain little from averaging \cite{fort_2019,neyshabur_2020,mustafa_2020}; we quantify this for formula-pretrained checkpoints with the participation ratio of the residual correlation matrix and evaluate the specific remedies used in practice. Frozen features from foundation vision transformers carry microstructural information usable by linear models \cite{whitman_2025_10,oquab_2023_dinov2}; we use such a member both as a decorrelator and as a baseline. The homogenization formulas used here are classical: the Maxwell Garnett expression for dilute inclusions \cite{garnett_1904} and the Halpin--Tsai equations for short fibers \cite{halpin_1976,tucker_1999_8}.

Averaging weights along the optimization trajectory is a standard tool: Polyak--Ruppert averaging of the iterates \cite{polyak_1992}, stochastic weight averaging of periodic snapshots \cite{izmailov_2018_swa}, and the decaying-weight exponential moving average (EMA), whose dynamics and benefits are characterized in recent work \cite{morales_2024_ema}. The bias correction that Adam applies to its running moments \cite{kingma_2015_adam} is the analogue of what our fixed EMA omits; since the average starts from the initial weights $\theta_0$ rather than zero, that correction is usually supplied instead by a decay warm-up, and without it the average stays close to $\theta_0$ early in training. A per-step EMA is a common default (training libraries such as timm expose it with an optional warm-up), so the confound we isolate is this default combined with a small step budget, not an unusual setting.

\section{Protocol}
\label{sec:protocol}
\subsection{Public benchmarks}
Table~\ref{tab:setup} lists the two benchmarks. Conductivity uses the two-phase RVEs of Li{\ss}ner and Fritzen \cite{lissner_2019,lissner_darus}: $400\times400$ binary images with circular or rectangular inclusions ($\kappa_m=1$, $\kappa_i=0.2$) and a stored effective-conductivity tensor from a Fourier-accelerated nodal solver, of which we regress $\kappa_{11}$; the fixed test set of 1,000 circular and 1,000 rectangular images was drawn once at seed 0. Stiffness uses the official training (4,000) and validation (1,000) RVEs of Wei and Chen \cite{wei_2024_11,wei_zenodo}, with the official validation set as our test set and $C_{1111}$ (GPa) as the target. Both benchmarks are simulated; ``real'' below denotes their images and labels, as opposed to our synthetic pretraining data. Real labels are drawn from the training pools at 100, 400, and 1,600 images; the 400-label budget has three pre-specified subsets (seeds 0, 1, 2) and is the primary comparison, while 100 and 1,600 use subset 0 only. All images are resized to $256\times256$. A third pretext, Al--Si die-casting hardness \cite{heo_2026}, was trained but is not evaluated here (Appendix~\ref{app:pretrain}). The networks fit the two formulas to $2.9\%$ and $1.3\%$ of the synthetic hold-out label standard deviation (Appendix~\ref{app:pretrain}), so the transfer differences below do not stem from underfitting the pretext.

\begin{table}[t]
\begin{center}
\footnotesize
\setlength{\tabcolsep}{4pt}
\resizebox{\linewidth}{!}{%
\begin{tabular}{|l|c|c|}
\hline
 & Conductivity & Stiffness \\
\hline\hline
Target & Solver $\kappa_{11}$ & FEM $C_{1111}$ \\
Source & Li{\ss}ner, DaRUS & Wei \& Chen, Zenodo \\
Real images & 4,795 & 5,000 \\
Test set & 2,000 (fixed) & 1,000 (official) \\
Label budgets & 100 / 400 / 1,600 & 100 / 400 / 1,600 \\
Synthetic images & 150,000 & 150,000 \\
Pretext formula & 2-D MG, Eq.~(\ref{eq:mg}) & HT surrogate, Eq.~(\ref{eq:ht}) \\
Synthetic label s.d. & 0.132 & 33.4 \\
Synthetic hold-out RMSE & 0.0038 & 0.440 \\
\hline
\end{tabular}}
\end{center}
\caption{Benchmarks and pretext tasks (MG: Maxwell Garnett; HT: Halpin--Tsai). The synthetic hold-out RMSE is the last-epoch error on 1,000 synthetic images withheld from pretraining; it monitors whether the formula was learned and is not a test metric. The label s.d.\ is over all 150,000 synthetic images; on the hold-out it is $0.130$ and $33.3$.}
\label{tab:setup}
\end{table}

\subsection{Synthetic generators and formula labels}
\begin{figure}[t]
\begin{center}
\includegraphics[width=0.76\linewidth]{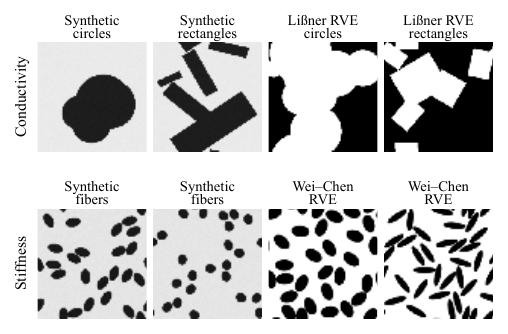}
\end{center}
\caption{Synthetic pretext images (left two columns) and public benchmark images (right two columns). For conductivity the generator renders the low-conductivity inclusions dark on a light matrix, whereas the public RVEs store them bright on a dark matrix; for stiffness the polarity agrees. The mismatch was found after all pre-specified results were computed.}
\label{fig:examples}
\end{figure}

The conductivity generator draws overlapping circles or rotated rectangles until a target inclusion fraction $f\in[0.18,0.84]$ is reached. Its label is the area-weighted two-dimensional Maxwell Garnett conductivity along the first axis,
\begin{equation}
\frac{\kappa_{11}}{\kappa_m}=\frac{1+f\beta(1-L)}{1-f\beta L},\qquad
\beta=\frac{\kappa_i-\kappa_m}{\kappa_m+L(\kappa_i-\kappa_m)},
\label{eq:mg}
\end{equation}
with depolarization factor $L=1/2$ for circles. A rectangle of aspect ratio $a\in[1,4]$ at angle $\theta$ is treated as an ellipse with $L_{\parallel}=1/(a+1)$ and $L_{\perp}=a/(a+1)$ and contributes $\kappa_{\parallel}\cos^2\theta+\kappa_{\perp}\sin^2\theta$, each $\kappa$ evaluated at the total fraction $f$; the image label is the area-weighted mean over its inclusions. The stiffness generator places 15 to 35 non-overlapping elliptical fibers with aspect ratio 1 to 4 and volume fraction $0.10$ to $0.40$, matching the descriptor ranges of the public set. Its label is an orientation-averaged Halpin--Tsai surrogate,
\begin{equation}
\begin{split}
E_x&=\langle\cos^2\theta\rangle E_L+\langle\sin^2\theta\rangle E_T,\\
C_{1111}&\approx\frac{E_x(1-\nu)}{(1+\nu)(1-2\nu)},
\end{split}
\label{eq:ht}
\end{equation}
where $E_L$ and $E_T$ follow the Halpin--Tsai form with shape parameters $\xi=2l/d$ and $\xi=2$, matrix and fiber moduli of 100 and 500\,GPa, and $\nu=0.30$. Neither formula was fit to any public label, and both are deliberately simple: Eq.~(\ref{eq:mg}) is a dilute approximation applied up to $f=0.84$, and Eq.~(\ref{eq:ht}) an isotropic surrogate of an anisotropic component. Fig.~\ref{fig:examples} shows synthetic and real images.

\subsection{Pretraining and fine-tuning}
The backbone is ConvNeXt-Small \cite{liu_2022_convnext} initialized from ImageNet weights. Formula pretraining regresses the standardized label with Smooth-L1 loss (transition point $0.5$), AdamW \cite{loshchilov_2019}, learning rate $3\times10^{-4}$, weight decay $0.05$, batch 48, cosine schedule, and 12 epochs over 150,000 images. The self-supervised control uses the same encoder, the same images, and the same schedule, but reconstructs $256\times256$ images from inputs with 60\% of $16\times16$ patches masked, following the masked-autoencoding paradigm \cite{he_2022_mae}: the dense encoder sees the masked image, a $1\times1$ convolution and bilinear upsampling form the decoder, the loss is the L1 error on masked pixels, and only the encoder is retained. The third initialization, ImageNet, omits the synthetic stage entirely. All images are grayscale, replicated to three channels, scaled to $[0,1]$, and normalized with the ImageNet mean and standard deviation; no per-dataset normalization or inversion is applied, so each image enters the network with its stored polarity. Fine-tuning is identical for all three: 30 epochs, learning rate $5\times10^{-5}$ (one sixth of the pretraining rate), head learning rate five times larger, weight decay $0.08$, dropout $0.1$, mixup $0.3$ \cite{zhang_2018_mixup}, gradient clipping at $1.0$, an EMA of weights with decay $0.995$, horizontal and vertical flips, and four-flip test-time averaging. Rotations by $90^\circ$ are not used because they exchange $\kappa_{11}$ and $\kappa_{22}$ (or $C_{1111}$ and $C_{2222}$). The EMA starts at the initial weights without bias correction, and its last value is used; no epoch is selected on validation or test data. The formula checkpoint keeps its pretrained regression head, whereas the ImageNet and masked-reconstruction initializations start with a new head. At 100 and 400 labels (90 and 360 steps), $64\%$ and $16\%$ of the EMA still rests on the initial weights, so these budgets are rechecked post hoc with the EMA disabled (Section~\ref{sec:ema}).

\subsection{Ensemble rules, DINOv2 member, and metrics}
For the primary budget (400 labels, subset 0), five fine-tuning seeds start from the formula checkpoint. We report each seed, the equal-weight mean $\bar y$, the pre-specified mixture
\begin{equation}
\hat y = 0.95\,\bar y + 0.05\,y_{\mathrm{DINOv2}},
\label{eq:mix}
\end{equation}
and a post-processing rule, fixed in advance, that shifts $\hat y$ to the training-label mean and multiplies deviations from it by $1.06$, using training predictions only. The DINOv2 member is a ridge regressor ($\alpha=50$) on standardized frozen features of \texttt{vit\_small\_patch14\_dinov2} at $252\times252$ (class token, patch mean, patch standard deviation) \cite{oquab_2023_dinov2}. Member dependence is summarized by the residual correlation matrix $R$ of the five seeds and its participation ratio
\begin{equation}
N_{\mathrm{eff}}=\frac{\left(\sum_j\lambda_j\right)^2}{\sum_j\lambda_j^2},
\label{eq:neff}
\end{equation}
where $\lambda_j$ are the eigenvalues of $R$; $N_{\mathrm{eff}}=5$ means independent members and $N_{\mathrm{eff}}=1$ means one effective member. The role of $\bar\rho$ follows from the mean squared error of an equal-weight average of $M$ members whose residuals have common variance $\sigma^2$ and mean pairwise correlation $\bar\rho$,
\begin{equation}
\mathbb{E}\big[(\bar y-y)^2\big]=\sigma^2\Big[\underbrace{\bar\rho}_{\text{shared}}+\underbrace{\tfrac{1-\bar\rho}{M}}_{\text{averaged out}}\Big].
\label{eq:shared}
\end{equation}
Equation~\eqref{eq:shared} separates the two effects studied here: a pretext label can lower the single-member error $\sigma^2$ (transfer), and a shared checkpoint can drive $\bar\rho$ toward one for its fine-tuned members (collapse); only the second term is removed by averaging, and at $\bar\rho=0.994$, $M=5$ it is $0.1\%$ of the total. Metrics are RMSE (primary), mean absolute error (MAE), and $R^2$ in original units. Uncertainty over label subsets is the standard deviation of RMSE across the three 400-label subsets; uncertainty over test images is a 95\% paired-bootstrap interval of the RMSE difference (4,000 resamples). The budget curves, ensemble rules, mixture weight, and ridge penalty were pre-specified; all other analyses, including the descriptor analysis of Section~\ref{sec:formula}, the polarity ablation and its follow-up runs, the ViT-Small (ViT-S) replication, and the EMA-free recheck, were designed after the test results were seen and are labeled post hoc. For each follow-up run, the settings and the decision rule were written to an internal protocol file before the run (Appendix~\ref{app:followup}).

\section{Results}
\subsection{Seed dependence under a shared formula checkpoint}
\label{sec:collapse}
Under the pre-specified EMA, the five formula-pretrained seeds are almost perfectly correlated (Fig.~\ref{fig:correlation}, Appendix~\ref{app:figs}). On conductivity the smallest off-diagonal entry is $0.986$ and the mean is $\bar\rho=0.994$; on stiffness the smallest entry is $0.9987$ and the mean is $0.999$. The participation ratio is $N_{\mathrm{eff}}=1.01$ and $1.00$: five networks that differ only in fine-tuning seed carry the information of one, regardless of the spread in single-seed accuracy ($0.0517$ to $0.0711$ on conductivity).
The consequence follows from Eq.~\eqref{eq:shared}: at $\bar\rho=0.994$ the five-seed average retains $99.5\%$ of a single member's mean squared error, and the ensemble RMSE nearly equals the mean of the member RMSEs ($0.0603$ versus $0.0605$ on conductivity, $4.774$ versus $4.775$ on stiffness). We treat $N_{\mathrm{eff}}$ as a summary of $\bar\rho$, not as independent evidence, since the two are nearly monotone for uniform correlations. The EMA-free recheck (post hoc; Section~\ref{sec:ema}) removes the reading of Fig.~\ref{fig:correlation} as a checkpoint effect. On conductivity $\bar\rho$ falls to $0.734$ ($N_{\mathrm{eff}}=1.55$) and the equal mean ($0.0108$) beats the best seed ($0.0113$), so the collapse there stems largely from the initial weights that all seeds share inside the EMA. On stiffness five formula seeds still have $\bar\rho=0.998$, but five ImageNet-initialized seeds reach $0.992$ and formula and ImageNet seeds correlate at $0.974$ with each other; without the ten largest formula residuals the formula and ImageNet values are $0.989$ and $0.952$. The stiffness collapse is therefore an error structure shared by the task, slightly tighter for the formula checkpoint, not a property of formula pretraining.

With the EMA weights the pre-specified ensemble rules gain little (Table~\ref{tab:rules}, Appendix~\ref{app:crossinit}): on conductivity the equal mean ($0.0603$), the $0.95/0.05$ DINOv2 mixture ($0.0576$), and the mean-alignment post-processing ($0.0526$) are all worse than the best seed ($0.0517$); on stiffness the equal mean and the mixture are within $0.012$ of the best seed ($4.763$), below the subset standard deviation ($0.065$).

\subsection{Transfer relative to ImageNet}
\label{sec:transfer}
Table~\ref{tab:curve} reports the label-budget curves for the three initializations, with one fine-tuning seed per cell (plotted in Fig.~\ref{fig:curves}, Appendix~\ref{app:figs}). The two materials differ in the sign of the effect.

\begin{table}[t]
\begin{center}
\footnotesize
\setlength{\tabcolsep}{3pt}
\resizebox{\linewidth}{!}{%
\begin{tabular}{|l|c|c|c|}
\hline
Labels & ImageNet & Masked reconstruction & Formula \\
\hline\hline
\multicolumn{4}{|l|}{Conductivity $\kappa_{11}$ (test s.d.\ 0.140)} \\
\hline
100 & \textbf{0.0723} & 0.7413 & 0.1882 \\
400 (3 subsets) & $\mathbf{0.0133\pm0.0010}$ & $0.0503\pm0.0027$ & $0.0528\pm0.0009$ \\
1,600 & \textbf{0.0041} & 0.0064 & 0.0055 \\
100, no EMA$^{\ast}$ & \textbf{0.0169} & 0.0540 & 0.0355 \\
400, no EMA$^{\ast}$ & \textbf{0.0090} & 0.0150 & 0.0119 \\
\hline
\multicolumn{4}{|l|}{Stiffness $C_{1111}$ (test s.d.\ 22.1)} \\
\hline
100 & 13.28 & 24.63 & \textbf{5.49} \\
400 (3 subsets) & $4.84\pm0.05$ & $6.75\pm0.44$ & $\mathbf{4.78\pm0.07}$ \\
1,600 & 4.49 & \textbf{4.48} & 4.49 \\
100, no EMA$^{\ast}$ & 5.20 & 5.50 & \textbf{4.57} \\
400, no EMA$^{\ast}$ & 4.53 & 4.57 & \textbf{4.52} \\
\hline
\end{tabular}}
\end{center}
\caption{Test RMSE by label budget (fine-tuning seed 0). The 400-label row is the mean and standard deviation over three pre-specified label subsets; the other rows use subset 0. The $\pm$ reflects label-subset variation only; for the conductivity formula cell, five seeds on subset 0 span $0.0517$--$0.0711$ (Table~\ref{tab:rules}). $^{\ast}$Post hoc, subset 0, last weights without the EMA (Section~\ref{sec:ema}). Bold marks the lowest RMSE in each row.}
\label{tab:curve}
\end{table}

\textbf{Stiffness.} With 100 labels the formula initialization reaches RMSE $5.49$ ($R^2=0.938$), against $13.28$ for ImageNet ($R^2=0.640$) and $24.63$ for masked reconstruction ($R^2<0$); the paired bootstrap interval of the formula-minus-ImageNet difference is $[-8.9,-6.7]$. Without the EMA the advantage shrinks but persists: over five fine-tuning seeds on subset 0 the formula initialization reaches $4.56$--$4.62$ against $5.19$--$6.09$ for ImageNet. Because the formula checkpoint keeps its pretrained head, we also fine-tuned it with a re-initialized head (post hoc; Appendix~\ref{app:ema}); this arm reaches $4.57$--$4.69$, so the advantage resides in the encoder rather than in a calibrated head. At 400 labels the two are tied with the EMA ($4.78\pm0.07$ versus $4.84\pm0.05$; interval $[-0.08,+0.10]$ on subset 0); without it five seeds give separated ranges ($4.49$--$4.52$ versus $4.53$--$4.64$), a $1.4\%$ difference. This separation rests on a post hoc rule (defined for the conductivity block, applied to stiffness after its ranges were seen) that overrides the sign-flip rule of Appendix~\ref{app:ema}, under which the single-seed difference ($+0.017$ with the EMA to $-0.019$ without) would read as EMA-dependent. The masked-reconstruction control is weak: its error exceeds ImageNet at 100 and 400 labels on both materials, with and without the EMA, so the pretext degraded the ImageNet features (Section~\ref{sec:discussion}). At 1,600 labels all three reach RMSE $4.48$--$4.49$ (MAE $0.92$--$1.03$); nine of the ten largest residuals are shared across models, and removing those ten lowers the formula and ImageNet RMSEs to $1.6$, so the plateau is set by a few shared outliers (post hoc; Appendix~\ref{app:outliers}).

\textbf{Conductivity.} ImageNet beats both pre-specified pretexts at every budget, with and without the EMA. At 100 labels the formula ($0.188$) avoids the masked-reconstruction failure ($0.741$, $R^2=-27$) but has $2.6$ times the ImageNet RMSE ($0.0723$); without the EMA the three errors are $0.0355$, $0.0540$, and $0.0169$. At 400 labels the formula ($0.0528\pm0.0009$) and masked reconstruction ($0.0503\pm0.0027$) tie, each about four times ImageNet ($0.0133\pm0.0010$); without the EMA the factor falls to $1.4$ for the formula ($0.0119$ versus $0.0084$, over three subsets and five seeds) and $1.7$ for masked reconstruction on subset 0 ($0.0150$ versus $0.0090$). At 1,600 labels, where the EMA retains a negligible $0.05\%$ of the initial weights, the gap is $0.0055$ versus $0.0041$.

\textbf{Formula label versus label-free pretext.} Reading Table~\ref{tab:curve} by columns, the formula label outperforms masked reconstruction of the same synthetic images on both materials at 100 labels, by factors of $3.9$ and $4.5$ with the EMA and $1.5$ and $1.2$ without it (intervals $[-0.020,-0.017]$ and $[-1.29,-0.68]$). This column comparison does not isolate the encoder: masked reconstruction, like ImageNet, starts fine-tuning with a new head, whereas the formula checkpoint keeps its pretrained head. The head was re-initialized as a control only for the stiffness comparison against ImageNet (Section~\ref{sec:ema} and Appendix~\ref{app:ema}, block B), where loading the encoder alone still beat ImageNet. Whether any synthetic stage outperforms omitting it differs between the materials: at 400 labels masked reconstruction exceeds the ImageNet error by a factor of $3.8$ on conductivity but $1.4$ on stiffness ($1.7$ and $1.01$ without the EMA), and the formula stage shares the conductivity penalty. In the post hoc ViT-S replication (Appendix~\ref{app:vit}; EMA weights), the ordering is preserved on stiffness ($5.38$ versus $16.14$ at 100 labels) and reversed on conductivity ($0.240$ versus $0.113$); correcting the polarity restores it only with the EMA (Section~\ref{sec:polarity}).

\subsection{EMA-free recheck (post hoc)}
\label{sec:ema}
The EMA updates a shadow copy $\bar\theta\leftarrow0.995\,\bar\theta+0.005\,\theta$ after each step, starting at the initial weights $\theta_0$, so after $T$ steps $\theta_0$ still carries weight $0.995^{T}$: $0.64$, $0.16$, and $0.0005$ at 100, 400, and 1,600 labels ($T=90$, $360$, $1{,}500$). The retained part differs between initializations, because the formula checkpoint keeps its pretrained head while the others start with a new one. We therefore refit seven cells (three stiffness and four conductivity initializations, including the polarity-matched checkpoint of Section~\ref{sec:polarity}) at 100 and 400 labels on subset 0 with the EMA disabled and everything else unchanged, and in later stages ran five fine-tuning seeds per cell across the three subsets. The settings and decision rules were fixed before each run (Appendix~\ref{app:ema}).

\begin{table}[t]
\begin{center}
\footnotesize
\setlength{\tabcolsep}{2.5pt}
\resizebox{\linewidth}{!}{%
\begin{tabular}{|l|cc|cc|cc|}
\hline
 & \multicolumn{2}{c|}{100 labels} & \multicolumn{2}{c|}{400 labels} & \multicolumn{2}{c|}{Slope, 400} \\
Initialization & EMA & none & EMA & none & EMA & none \\
\hline\hline
\multicolumn{7}{|l|}{Conductivity $\kappa_{11}$} \\
\hline
ImageNet & 0.0723 & \textbf{0.0169} & 0.0126 & 0.0090 & 0.96 & 0.98 \\
Masked reconstruction & 0.7413 & 0.0540 & 0.0532 & 0.0150 & 0.98 & 1.01 \\
Formula, mismatched & 0.1882 & 0.0355 & 0.0517 & 0.0119 & 0.64 & 0.97 \\
Formula, matched$^{\dagger}$ & 0.0351 & 0.0223 & 0.0190 & \textbf{0.0085} & 1.03 & 0.99 \\
\hline
\multicolumn{7}{|l|}{Stiffness $C_{1111}$} \\
\hline
ImageNet & 13.28 & 5.20 & 4.78 & 4.53 & 0.96 & 0.97 \\
Masked reconstruction & 24.63 & 5.50 & 7.38 & 4.57 & 1.18 & 0.97 \\
Formula & 5.49 & \textbf{4.57} & 4.79 & \textbf{4.52} & 1.00 & 0.98 \\
\hline
\multicolumn{7}{|l|}{Five fine-tuning seeds at 400 labels: $\bar\rho$ with EMA $\rightarrow$ none} \\
\hline
Stiffness formula & \multicolumn{6}{c|}{$0.999\rightarrow0.998$} \\
Stiffness ImageNet & \multicolumn{6}{c|}{not run $\rightarrow0.992$} \\
Cond.\ formula, mismatched & \multicolumn{6}{c|}{$0.994\rightarrow0.734$} \\
Cond.\ formula, matched$^{\dagger}$ & \multicolumn{6}{c|}{$0.958\rightarrow0.873$} \\
\hline
\end{tabular}}
\end{center}
\caption{Post hoc EMA-free recheck (subset 0, fine-tuning seed 0): test RMSE and slope of prediction on truth with the pre-specified EMA weights and with the last weights. $^{\dagger}$Post hoc polarity-matched checkpoint. Bold marks the lowest RMSE per material and column without the EMA.}
\label{tab:ema}
\end{table}

Table~\ref{tab:ema} gives the outcome. The EMA increased every error, most where the initial weights are farthest from the target: masked reconstruction on conductivity (factor $14$ at 100 labels) and the mismatched formula checkpoint (factors $5.3$ and $4.3$), whose retained weights carry the inverted readout. Without the EMA the mismatched slope is $0.97$ not $0.64$ and the residual--descriptor correlation falls from $+0.89$ to $+0.32$, so the compression resides in the retained initial weights, not the fine-tuned network. Under rules fixed before the recheck, three conclusions survive without the EMA: formula beats ImageNet at 100 stiffness labels, ImageNet beats the mismatched formula at 100 and 400 conductivity labels, and the stiffness seeds collapse, though ImageNet seeds collapse too (Section~\ref{sec:collapse}). Three change: the matched formula, better than ImageNet at 100 conductivity labels with the EMA, is worse without it; at 400 labels the polarity comparison is inconclusive over three runs per side (Section~\ref{sec:polarity}, Appendix~\ref{app:mismatched}); and the conductivity seed collapse weakens. The 100-label conductivity cells use one subset and one seed; the ensemble rules, cross-checkpoint correlations, mixed-polarity runs, and stiffness ViT-S results use the EMA only.

These multi-subset, multi-seed results are collected in Table~\ref{tab:ema_final}, our most statistically robust comparison (three label subsets and five fine-tuning seeds): we read the small-budget conclusions from it and treat the pre-specified EMA budget curves of Table~\ref{tab:curve} as the pre-specified baseline they recheck. Block~(A) compares three matched against three mismatched pretraining runs (the mismatched side completed post hoc; Appendix~\ref{app:mismatched}): the matched run means ($0.0086$/$0.0111$/$0.0118$) separate below the mismatched means ($0.0119$/$0.0120$/$0.0123$) in aggregate but overlap on two of three subsets, so the comparison is inconclusive once the EMA is off, with a consistent but underpowered direction and a benefit much more variable on the matched side. Block~(B) shows both stiffness formula arms, head kept ($4.559$--$4.616$) or re-initialized ($4.567$--$4.687$), below ImageNet ($5.187$--$6.087$) at 100 labels, so the advantage is in the encoder; block~(C) that with a ViT-S backbone and matched polarity the formula label beats ViT ImageNet but trails ViT-MAE. Decision rules, all fixed before the runs, are in Appendix~\ref{app:ema}.

\begin{table}[t]
\begin{center}
\footnotesize
\setlength{\tabcolsep}{3pt}
\resizebox{\linewidth}{!}{%
\begin{tabular}{|l|c|c|c|c|}
\hline
 & Subset 0 & Subset 1 & Subset 2 & Seed range \\
\hline\hline
\multicolumn{5}{|l|}{(A) ConvNeXt, conductivity, 400 labels: mean of five seeds} \\
\hline
Formula, mismatched, run 0 & 0.0121 & 0.0123 & 0.0114 & 0.0101--0.0137 \\
Formula, mismatched, run 1 & 0.0118 & 0.0120 & 0.0123 & 0.0099--0.0173 \\
Formula, mismatched, run 2 & 0.0110 & 0.0133 & 0.0127 & 0.0100--0.0146 \\
Formula, matched, run 0 & 0.0083 & 0.0088 & 0.0086 & 0.0077--0.0094 \\
Formula, matched, run 1 & 0.0106 & 0.0116 & 0.0112 & 0.0101--0.0122 \\
Formula, matched, run 2 & 0.0115 & 0.0119 & 0.0120 & 0.0106--0.0129 \\
ImageNet & 0.0083 & 0.0086 & 0.0082 & 0.0069--0.0104 \\
\hline
\multicolumn{5}{|l|}{(B) ConvNeXt, stiffness, 100 labels, subset 0: mean of five seeds} \\
\hline
Formula, head kept & \multicolumn{3}{c|}{4.577} & 4.559--4.616 \\
Formula, head re-initialized & \multicolumn{3}{c|}{4.604} & 4.567--4.687 \\
ImageNet & \multicolumn{3}{c|}{5.554} & 5.187--6.087 \\
\hline
\multicolumn{5}{|l|}{(C) ViT-S, conductivity, 400 labels, matched polarity: seed 0} \\
\hline
Formula & 0.0095 & 0.0088 & 0.0091 & -- \\
ViT-MAE & 0.0061 & 0.0074 & 0.0077 & -- \\
ImageNet & 0.0151 & 0.0106 & 0.0156 & -- \\
\hline
\end{tabular}}
\end{center}
\caption{Primary EMA-free results (post hoc; test RMSE with the last weights). Over three label subsets and five fine-tuning seeds, this is the basis for the small-budget conclusions. In block~(A) both polarities now have three pretraining runs: matched runs 0--2 are the inverted pretrainings of Section~\ref{sec:polarity}, and mismatched runs 1--2 repeat the original (non-inverted) pretraining with new seeds (post hoc; Appendix~\ref{app:mismatched}). Under the pre-specified run-summary rule the verdict is \emph{not separated} (inconclusive): the matched run means (0.0086/0.0111/0.0118) separate below the mismatched means (0.0119/0.0120/0.0123) in aggregate but their per-subset ranges overlap on two of three subsets, and the benefit is heterogeneous across matched runs. Seed ranges are pooled over subsets. Per-block decision rules are in Appendix~\ref{app:ema}.}
\label{tab:ema_final}
\end{table}

\subsection{Formula error and transfer (post hoc)}
\label{sec:formula}
Our prior expectation was that transfer would track how close the pretext formula is to the real property. This post hoc analysis uses no training: each formula is evaluated on descriptors of the real test images --- the inclusion area fraction from each conductivity image (the bright phase, correlating at $-0.97$ with $\kappa_{11}$; Eq.~\eqref{eq:mg} with $L=1/2$) and the tabulated volume fraction and aspect ratio of each stiffness RVE with isotropic orientation. By RMSE the expectation fails (Table~\ref{tab:formula}, Fig.~\ref{fig:formula}): the conductivity formula lies within $0.32$ label standard deviations of the truth and the stiffness surrogate within $1.45$, yet the stiffness network adapts while the conductivity network has four times the ImageNet error at 400 labels ($1.4$ times without the EMA). By correlation with the truth the two formulas are equally accurate ($0.971$ and $0.976$) and a head absorbs affine error, so formula accuracy does not separate the materials and the transfer difference originates elsewhere.

\begin{table}[t]
\begin{center}
\footnotesize
\setlength{\tabcolsep}{4pt}
\resizebox{\linewidth}{!}{%
\begin{tabular}{|l|c|c|}
\hline
Predictor at 400 labels (subset 0) & Conductivity & Stiffness \\
\hline\hline
Formula on real descriptors, zero-shot & 0.0442 & 32.06 \\
\quad Pearson $r$ with truth & 0.971 & 0.976 \\
\quad Slope on truth & 0.92 & 1.37 \\
\quad Mean bias & $+0.0289$ & $+30.26$ \\
\quad RMSE / label s.d. & 0.32 & 1.45 \\
Formula + affine fit on 400 labels & 0.0334 & 4.843 \\
Formula-pretrained CNN, seed 0 & 0.0517 & 4.795 \\
ImageNet CNN & 0.0126 & 4.778 \\
DINOv2 frozen + ridge & 0.0242 & 4.951 \\
\hline
\multicolumn{3}{|l|}{Diagnostics, EMA weights / last weights without EMA} \\
\hline
Slope on truth, formula CNN & 0.64 / 0.97 & 1.00 / 0.98 \\
Slope on truth, ImageNet CNN & 0.96 / 0.98 & 0.96 / 0.97 \\
Residual--descriptor corr., formula CNN & $+0.89$ / $+0.32$ & $0.00$ / $-0.09$ \\
Residual--descriptor corr., ImageNet CNN & $+0.34$ / $+0.16$ & $-0.15$ / $-0.15$ \\
\hline
\end{tabular}}
\end{center}
\caption{Test RMSE of formula-based baselines on measured or tabulated descriptors, compared with the fine-tuned networks, and diagnostics of the fine-tuned predictions. The affine fit uses only the 400 training labels of subset 0; the descriptor is the inclusion area fraction (conductivity) or the fiber volume fraction (stiffness). CNN: fine-tuned ConvNeXt-Small; RMSE rows use the pre-specified EMA weights.}
\label{tab:formula}
\end{table}

The two formula errors differ in structure. The stiffness surrogate is far from the truth in absolute terms (bias $+30$ units) but almost entirely an affine distortion ($r=0.976$, slope $1.37$) that a regression head corrects from few labels; a two-parameter affine fit alone reaches $4.843$. The conductivity formula is close to the truth with a nearly uniform bias ($+0.029$), yet its network transfers worse. With the EMA the failure mode is visible in Fig.~\ref{fig:formula}: predictions span $0.31$--$0.66$ against a truth range $0.25$--$0.78$ (slope $0.64$), with the residual rising monotonically in inclusion fraction ($r=0.89$) --- the averaged network has not inherited the formula's error but lost sensitivity to the quantity it depends on. This compression is absent for the ImageNet model (slope $0.96$), the stiffness formula model ($1.00$), and the same conductivity model without the EMA ($0.97$; Section~\ref{sec:ema}).

The affine baseline puts the failure in scale: the zero-shot formula ($0.0442$) and its calibration ($0.0334$) both beat every EMA-averaged mismatched-polarity ConvNeXt at 400 labels ($0.047$--$0.071$) but not the same checkpoint without the EMA ($0.0119$), so the fine-tuned network carries more than the affine signal, and every polarity-matched network beats the zero-shot formula, so the approximations of Eq.~\eqref{eq:mg} do not cap the fine-tuned error.

\subsection{A post hoc polarity ablation}
\label{sec:polarity}
Two observations locate the conductivity penalty on the image side. First, the penalty relative to ImageNet is much larger on conductivity than on stiffness for the label-free pretext as well (factors of $3.8$ and $1.4$), so part of it stems from the synthetic images rather than from the label. Second, Fig.~\ref{fig:examples} shows that the synthetic generator draws the low-conductivity phase dark whereas the public RVEs store it bright, while the stiffness polarities agree. A network that learned ``dark area fraction maps to $\kappa_{11}$'' starts fine-tuning with the wrong sign of that readout; fine-tuning reverses it (slope $0.97$ without the EMA), but the EMA keeps part of the pretrained weights, which appears as a compressed slope.

To test this explanation, we ran one post hoc pretraining after the pre-specified results were complete and fine-tuned it on the three 400-label subsets (seed 0) and with seeds 1--4 on subset 0: the 150,000 synthetic conductivity images were inverted, $x\mapsto255-x$, so that the low-conductivity inclusions are bright as in the public data, with everything else as in Section~\ref{sec:protocol}. The inverted images carry the intended phase meaning and are fitted as well as the originals (Appendices~\ref{app:pretrain} and~\ref{app:polarity}).

Table~\ref{tab:polarity} and Fig.~\ref{fig:polarity} (Appendix~\ref{app:polarity}) give the outcome; all entries use the EMA weights. With matched polarity the three-subset RMSE falls from $0.0528\pm0.0009$ to $0.0170\pm0.0014$, the slope rises to $1.01$--$1.03$ (from $0.63$--$0.64$), and the residual--descriptor correlation turns negative ($-0.30$ to $-0.45$, from $+0.89$--$+0.90$), so the compressed readout of Section~\ref{sec:formula} requires both the mismatched polarity and the EMA. With the EMA the matched model beats the affine-calibrated formula ($0.0334$) and the DINOv2 ridge ($0.0238$) but keeps $1.28$ times the ImageNet RMSE. Without the EMA (Appendix~\ref{app:ema}) the first matched run reaches $0.0086$, overlapping ImageNet ($0.0084$) and below the mismatched run ($0.0119$), but two further matched runs reach $0.0111$ and $0.0118$; completing the mismatched side to three runs as well ($0.0119$--$0.0123$; Appendix~\ref{app:mismatched}) leaves the comparison inconclusive and heterogeneous (Section~\ref{sec:discussion}), with no matched seed range below ImageNet. The matched seeds have $N_{\mathrm{eff}}=1.07$ with the EMA and $1.23$ without. Fine-tuning the original checkpoint with 50\% random inversion is harmful (RMSE $0.110$, slope $0.37$): the same label then appears under both polarities, so this measures how hard a polarity-invariant readout is to learn from 400 labels, not where the shortcut resides.

\begin{table}[t]
\begin{center}
\footnotesize
\setlength{\tabcolsep}{3pt}
\resizebox{\linewidth}{!}{%
\begin{tabular}{|l|c|c|c|c|}
\hline
Initialization & 100 & 400 (3 subsets) & 1,600 & Slope \\
\hline\hline
\multicolumn{5}{|l|}{ConvNeXt-Small} \\
\hline
Formula, mismatched (pre-spec.) & 0.1882 & $0.0528\pm0.0009$ & 0.0055 & 0.63--0.64 \\
Formula, matched & \textbf{0.0351} & $0.0170\pm0.0014$ & 0.0053 & 1.01--1.03 \\
\quad pretraining run 1 & -- & $0.0195\pm0.0009$ & -- & 0.97--0.99 \\
\quad pretraining run 2 & -- & $0.0208\pm0.0002$ & -- & 0.98 \\
Formula, mixed polarity & 0.0353 & $0.0192\pm0.0004$ & 0.0057 & 0.99--1.00 \\
Formula, mismatched + inv.\ aug. & -- & 0.1104 (subset 0) & -- & 0.37 \\
ImageNet (pre-spec.) & 0.0723 & $\mathbf{0.0133\pm0.0010}$ & \textbf{0.0041} & 0.94--0.96 \\
\hline
\multicolumn{5}{|l|}{ViT-S} \\
\hline
Formula, mismatched & 0.2402 & $0.0507\pm0.0023$ & 0.0057 & 0.70--0.75 \\
Formula, matched & \textbf{0.0351} & $0.0222\pm0.0015$ & 0.0050 & 0.98--0.99 \\
ViT-MAE, mismatched & 0.1125 & $0.0341\pm0.0005$ & 0.0041 & 0.76--0.77 \\
ViT-MAE, matched & 0.1125 & $0.0343\pm0.0015$ & \textbf{0.0039} & 0.76--0.78 \\
ImageNet & 0.0475 & $\mathbf{0.0190\pm0.0022}$ & 0.0052 & 0.92--0.96 \\
\hline
\end{tabular}}
\end{center}
\caption{Conductivity test RMSE of the polarity variants by label budget (fine-tuning seed 0; 400 labels: mean $\pm$ s.d.\ over three subsets; 100 and 1,600: subset 0). Slope: prediction on truth, range over the three 400-label subsets. Only the rows marked pre-spec.\ are pre-specified; all ViT-S rows are post hoc. Inv.\ aug.: 50\% random inversion during fine-tuning only; mixed polarity: 50\% random inversion during pretraining only. Bold marks the best entry per backbone and budget; --: not run. All entries use the EMA weights; values without the EMA are in Table~\ref{tab:ema} and Appendix~\ref{app:ema}.}
\label{tab:polarity}
\end{table}

\textbf{Follow-up runs.} Further post hoc runs, with settings and rules fixed in advance (Appendix~\ref{app:followup}), appear in Table~\ref{tab:polarity}. (i) Two more inverted pretrainings give $0.0195\pm0.0009$ and $0.0208\pm0.0002$, each below the mismatched and above ImageNet with the EMA but not consistently separated from the mismatched model without it. (ii) At 100 labels the matched checkpoint beats ImageNet with the EMA ($0.0351$ versus $0.0723$) but not without it ($0.0223$ versus $0.0169$), so we claim no 100-label advantage. (iii) With ViT-S and inverted images for both pretexts, the formula label beats the ViT masked autoencoder (ViT-MAE) on all three 400-label subsets with the EMA but loses on all three without it ($0.0088$--$0.0095$ versus $0.0061$--$0.0077$; one seed), while still beating ViT ImageNet ($0.0106$--$0.0156$). (iv) Mixed-polarity pretraining (each image inverted with probability $0.5$) needs no target-polarity knowledge; it lowers conductivity to $0.0192\pm0.0004$ and 400-label stiffness to $4.72\pm0.06$ but raises the 100-label stiffness RMSE ($5.49\to5.54$), failing its no-harm rule as a single-seed difference at the budget most affected by the EMA.

\subsection{Frozen foundation features as a baseline}
\label{sec:frozen}
Ridge regression on frozen DINOv2 features, with the penalty fixed in advance, reaches $0.0238\pm0.0002$ on conductivity and $4.91\pm0.03$ on stiffness over the three 400-label subsets. On conductivity it is about half any EMA-averaged mismatched-polarity network and $1.8$ times ImageNet fine-tuning, while every EMA-free ConvNeXt is better (at most $0.0150$); on stiffness it matches the EMA-averaged formula and ImageNet networks but trails all three EMA-free ones ($4.52$--$4.57$ against $4.95$ on subset 0). Needing one forward pass per image and no GPU training, it is a useful floor for any claimed benefit of physics pretraining, and its strength is the DINOv2 features, not the ridge: the same ridge on frozen ImageNet ConvNeXt-Small features reaches only $0.0521\pm0.0004$ and $5.37\pm0.08$ (post hoc; Appendix~\ref{app:imagenet}).

\section{Discussion}
\label{sec:discussion}
\textbf{Transfer and seed dependence.} A formula label was expected to give a property-specific initialization and, as a side effect, a shared error pattern among networks fine-tuned from one checkpoint. For ConvNeXt only the first survives the recheck, mainly on stiffness at the smallest budget: without the EMA, over five seeds, formula pretraining lowers the 100-label error by $18\%$ relative to ImageNet, also with a re-initialized head, and is within $1.5\%$ at 400 and 1,600 labels. On conductivity one of three matched runs reaches the ImageNet level, and none lies below it. The high stiffness seed correlation is shared by ImageNet seeds ($0.992$) and by formula--ImageNet pairs ($0.974$), so it is set by the task's common errors, and on conductivity most of the collapse came from the EMA. The polarity effect is large only with the EMA; without it the comparison is inconclusive and heterogeneous, as the next paragraph details.

\textbf{Pretraining-run variation is itself a result.} The three matched runs keep the same order (run $0<1<2$) with and without the EMA ($0.0170/0.0195/0.0208$ versus $0.0086/0.0111/0.0118$ at 400 labels), so the gaps are a real difference in checkpoint quality, not fine-tuning noise, and fine-tuning seeds do not capture it. The ordering also runs against the pretraining fit: run 0 has the worst synthetic hold-out RMSE ($0.0041$ versus $0.0038$ for runs 1 and 2) yet transfers best, so how well a checkpoint fit the formula pretext does not predict how well it transfers. A single pretraining run can therefore fall anywhere within this spread, which is why the polarity verdict needs several runs on each side. Comparing three matched against three mismatched runs without the EMA is inconclusive: the variation falls almost entirely on the matched side ($0.0086$--$0.0118$ versus $0.0119$--$0.0123$ for mismatched), so matching the polarity ranges from no gain to the ImageNet level across runs, and three runs give too little power to separate a heterogeneous effect (Appendix~\ref{app:mismatched}). We turn this into a checklist item below.

\textbf{Weight averaging at small budgets.} A per-step EMA without bias correction, a common default, returns after few steps a blend of the initial and fine-tuned weights. The blend penalizes every initialization unequally (least the formula checkpoint with its pretrained head, most an encoder far from the target or a retained wrong-sign readout) and gives all seeds the same initial component, raising their residual correlation. Because the schedule was fixed in epochs, 100 labels gave only 90 steps; a fixed step count, bias correction, or an EMA warm-up would likely have reduced the confound.

\textbf{Scope of the label-free control.} The pre-specified masked-reconstruction control degraded the ImageNet features with and without the EMA. The post hoc ViT masked autoencoder \cite{he_2022_mae} (Appendix~\ref{app:vit}) is stronger, with the lowest 1,600-label RMSE on both materials (one seed): with the EMA it beat the formula label on conductivity under mismatched polarity and fell behind with aligned images, and without the EMA it beats the aligned formula label on all three subsets (one seed).

\textbf{Checklist before pretraining on formula labels.}
\begin{enumerate}
\item \textbf{Calibrate the formula first.} Evaluate it zero-shot on descriptors of a few real images with a two-parameter affine fit. The calibrated conductivity formula ($0.0334$) beat every EMA-averaged mismatched network at 400 labels but none without the EMA; on stiffness it matched only the EMA-averaged ones ($4.843$ versus $4.78\pm0.07$).
\item \textbf{Verify the images against the target.} Compare polarity, contrast, and resolution. Matching the conductivity polarity cut the 400-label RMSE by a factor of $2.5$--$3.1$ with the EMA; without it the gain ranged from a factor of $1.4$ to none across three pretraining runs.
\item \textbf{Repeat the pretraining several times on each side, ideally at least five.} Within one polarity, pretraining runs of the same recipe spanned $0.0086$--$0.0118$ without the EMA at 400 conductivity labels, as large as the effect under test, and fine-tuning seeds did not reveal it. Three runs per side are too few: a one-sided run-level comparison then cannot fall below $p=1/20$, and our contrast test came out inconclusive at that floor. Report a run-level effect estimate with an interval, not a single checkpoint, and prefer five or more pretraining runs per condition.
\item \textbf{Check the weight average.} At small budgets report the last weights next to any averaged weights for every initialization; here the EMA inflated errors by up to a factor of $14$ and reversed two comparisons with ImageNet.
\item \textbf{Do not count seeds as members.} Without the EMA, five seeds gave $N_{\mathrm{eff}}=1.00$ (formula) and $1.01$ (ImageNet) on stiffness and $1.23$--$1.55$ on conductivity; measure $N_{\mathrm{eff}}$ against another initialization before weighting.
\item \textbf{Report a frozen-feature baseline.} Ridge on frozen DINOv2 features ($0.0238\pm0.0002$ and $4.91\pm0.03$) needs no training and bounds what pretraining must beat.
\end{enumerate}

\textbf{Limitations.} The pre-specified study uses one backbone (ViT-S is post hoc), one pretraining schedule, one split seed, three label subsets at the primary budget, and one fine-tuning seed per budget-curve cell; bootstrap intervals cover test-image sampling only, not training randomness, which on conductivity at 400 labels spans $0.0517$--$0.0711$ across seeds with the EMA and $0.0101$--$0.0137$ without it. Both conductivity polarities were pretrained three times for the EMA-free comparison, but the stiffness and ViT-S pretexts have a single pretraining run, and the 100-label conductivity and ViT-S cells have one fine-tuning seed. The ensemble rules, cross-checkpoint correlations, stiffness ViT-S replication, and mixed-polarity runs use the EMA only. The conductivity numbers confound a formula label with a misaligned generator, and the EMA-free polarity comparison across three runs per polarity is underpowered (a run-level test cannot fall below $p=1/20$) for the heterogeneous effect observed. The remaining synthetic-to-real gap was not decomposed into histogram, shape, and noise components.

\section{Conclusion}
One pre-specified protocol on two public benchmarks, with post hoc rechecks, supports three conclusions. First, the pre-specified weight average, which still held $64\%$ and $16\%$ of the initial weights at 100 and 400 labels, inflated small-budget errors, accounted for most of the conductivity seed collapse, and, with the inverted polarity of the conductivity generator, the compressed readout; small-budget comparisons of pretrained initializations should report the last weights as well. Second, without the EMA, the ConvNeXt formula label helps mainly at 100 stiffness labels, by $18\%$ over five seeds, via its encoder; on conductivity, matching polarity reaches ImageNet for one of three runs but, tested with three pretraining runs on each polarity, gives an inconclusive and heterogeneous effect (large on some runs, near zero on others), and in ViT-S the matched label beats ImageNet (one seed, with a wide subset spread of $0.0106$--$0.0156$ for ImageNet) but trails ViT-MAE. Third, the near-unit seed correlation on stiffness appears for ImageNet seeds as well, so it reflects errors shared by the task; seeds should not be counted as ensemble members without measuring $N_{\mathrm{eff}}$ against another initialization. Frozen DINOv2 features and an affine-calibrated formula beat the EMA-averaged mismatched-polarity ConvNeXt networks but not the EMA-free ones, and remain useful baselines. Generators, code, saved predictions, and checkpoints will be released.

\clearpage
\bibliography{references}

\appendix
\renewcommand\thesection{\Alph{section}}
\setcounter{table}{0}
\setcounter{figure}{0}
\renewcommand\thetable{A\arabic{table}}
\renewcommand\thefigure{A\arabic{figure}}

\section{Experimental Environment}
\label{app:env}
All experiments, pre-specified and post hoc, were run on a single workstation with one NVIDIA GeForce RTX 4080 SUPER (16\,GB), CUDA 12.4, Python 3.13.5, PyTorch 2.6.0, timm 1.0.26, scikit-learn 1.5.2, and NumPy 2.5.2. Training uses automatic mixed precision with a gradient scaler, AdamW, and a cosine schedule; the backbone is \texttt{convnext\_small.in12k\_ft\_in1k} from timm with global-average pooling and a linear head, and the frozen foundation features come from \texttt{vit\_small\_patch14\_dinov2.lvd142m} evaluated at $252\times252$ pixels (class token concatenated with the mean and standard deviation of the patch tokens, 1,152 dimensions). Synthetic generation, label splits, fine-tuning seeds, and the DINOv2 ridge are all seeded; the real-image splits are fixed by one split seed and stored with the cached datasets. Every pre-specified number in the paper is read from a single JSON file produced by the evaluation script; the post hoc analyses write their own JSON files to a separate directory and read the pre-specified predictions and checkpoints without modifying them. The first post hoc polarity run (inverted pretraining for 12 epochs on 150,000 images, two 30-epoch fine-tuning runs, and ImageNet feature extraction for both datasets) took 6,226\,s (104 minutes) on this machine; the replication (six further fine-tunings from the inverted checkpoint) took about six minutes, and the ViT-S replication of Appendix~\ref{app:vit} took 15,278\,s (4.2 hours). Of the follow-up runs (Appendix~\ref{app:followup}), the two further inverted pretrainings with their fine-tunings took 13,136\,s (3.6 hours) and the polarity-corrected ViT-S run 7,738\,s (2.1 hours); the wall time of the mixed-polarity run is not available. The formula seed stage of the EMA-free recheck (Appendix~\ref{app:ema}; twelve 400-label fine-tunings) took 655\,s and the ImageNet seed stage (four fine-tunings) 225\,s; the final stage took 3,477\,s for 64 new conductivity fine-tunings, 228\,s for the stiffness head control, and 335\,s for the ViT-S cells. The mismatched-polarity replication (Appendix~\ref{app:mismatched}) added two non-inverted pretrainings and their fine-tunings: pretraining seed 1 took 6,000\,s and its fine-tunings 818\,s, while pretraining seed 2 was interrupted at epoch 11 and resumed (558\,s for the resumed segment) with fine-tunings of 816\,s, for roughly 13,600\,s in total; the driver script reported only its final 1,375\,s invocation because it reused the cached seed-1 and resumed seed-2 checkpoints.

\section{Pretraining Diagnostics}
\label{app:pretrain}
\begin{figure}[t]
\begin{center}
\includegraphics[width=\linewidth]{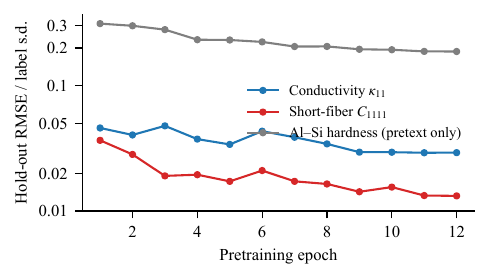}
\end{center}
\caption{Synthetic hold-out RMSE of formula pretraining, divided by the hold-out label standard deviation, per epoch. A third pretext, an aluminum--silicon hardness formula (gray), was trained but is not evaluated on public labels in this paper.}
\label{fig:pretrain}
\end{figure}

Fig.~\ref{fig:pretrain} shows the synthetic hold-out RMSE per epoch for the three formula pretexts, divided by the hold-out label standard deviation ($0.130$ for conductivity, $33.3$ for stiffness). Both evaluated pretexts converge without epoch selection; the hardness pretext (gray) is included for completeness and is not evaluated on public labels in this paper. Table~\ref{tab:pretrain_inv} lists the unnormalized hold-out RMSE of the inverted conductivity pretraining used in Section~\ref{sec:polarity}: the trajectory is slightly noisier in the first epochs but ends at $0.0041$, against $0.0038$ for the original, so the inverted images are fitted as well as the originals and the fine-tuning difference does not stem from pretraining fit.

\begin{table}[t]
\begin{center}
\footnotesize
\setlength{\tabcolsep}{3pt}
\resizebox{\linewidth}{!}{%
\begin{tabular}{|l|cccccc|}
\hline
Epoch & 1 & 2 & 3 & 4 & 5 & 6 \\
\hline
RMSE & 0.0055 & 0.0169 & 0.0071 & 0.0061 & 0.0048 & 0.0057 \\
\hline\hline
Epoch & 7 & 8 & 9 & 10 & 11 & 12 \\
\hline
RMSE & 0.0058 & 0.0049 & 0.0043 & 0.0042 & 0.0041 & 0.0041 \\
\hline
\end{tabular}}
\end{center}
\caption{Synthetic hold-out RMSE per epoch of the inverted-polarity conductivity pretraining (post hoc). Final value $0.0041$, i.e.\ $3.2\%$ of the hold-out label standard deviation; the original pretraining ends at $0.0038$ ($2.9\%$).}
\label{tab:pretrain_inv}
\end{table}

\section{Budget Curves and Formula Diagnostics}
\label{app:figs}
\begin{figure}[t]
\begin{center}
\includegraphics[width=0.78\linewidth]{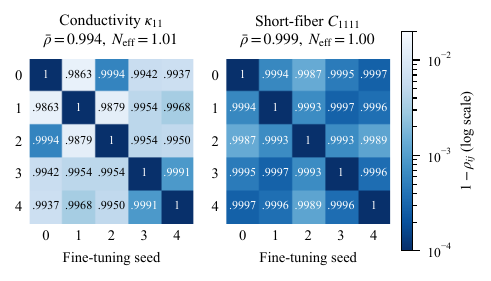}
\end{center}
\caption{Residual correlation $\rho_{ij}$ between five fine-tuning seeds that share one formula-pretrained checkpoint (400 labels, subset 0, pre-specified EMA weights); color encodes $1-\rho_{ij}$ on a log scale. The smallest off-diagonal entries are $0.986$ (conductivity) and $0.9987$ (stiffness).}
\label{fig:correlation}
\end{figure}

\begin{figure*}[t]
\begin{center}
\includegraphics[width=0.7\textwidth]{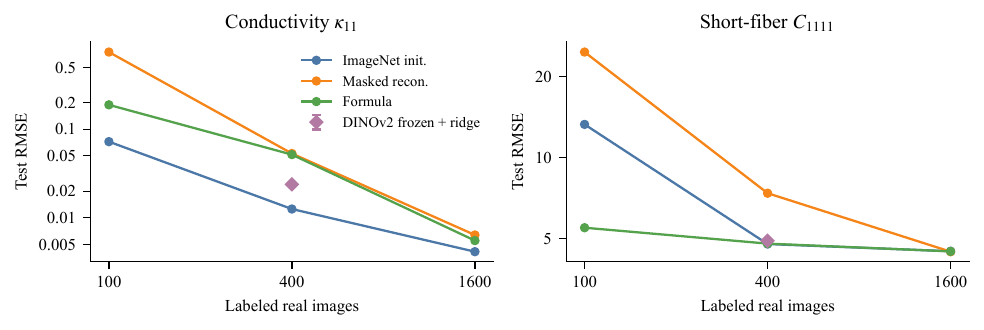}
\end{center}
\caption{Test RMSE against the number of real labels (subset 0, fine-tuning seed 0, EMA weights, log axes). The diamond is the frozen DINOv2 ridge model at 400 labels (mean and s.d.\ over three subsets).}
\label{fig:curves}
\end{figure*}

\begin{figure*}[t]
\begin{center}
\includegraphics[width=0.7\textwidth]{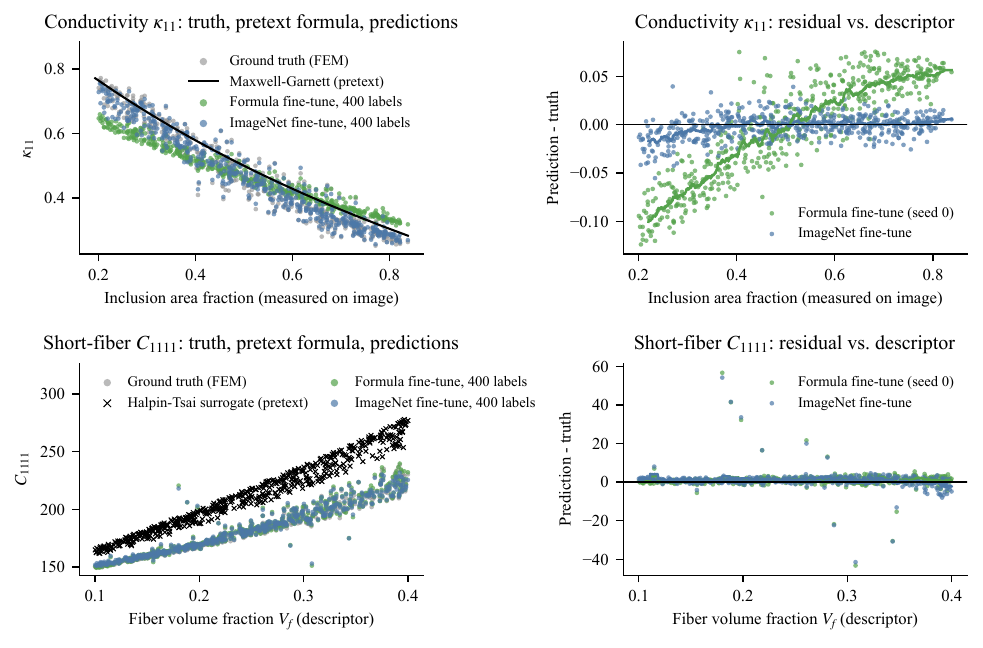}
\end{center}
\caption{Rows: conductivity (top) and stiffness (bottom). Left: ground truth (gray), the pretext formula evaluated on the real descriptor (black), and the 400-label predictions (EMA weights) of the formula-pretrained (green) and ImageNet-initialized (blue) networks. Right: residuals against the same descriptor with a moving average. The formula-pretrained conductivity model is compressed toward the mean, which the ImageNet model is not; on stiffness both models are unbiased across the volume-fraction range even though the pretext formula has a bias of $+30$ units.}
\label{fig:formula}
\end{figure*}

Fig.~\ref{fig:correlation} shows the seed residual correlations of Section~\ref{sec:collapse}, Fig.~\ref{fig:curves} plots the pre-specified budget curves of Table~\ref{tab:curve}, and Fig.~\ref{fig:formula} shows the descriptor-level diagnostics of Section~\ref{sec:formula}; all use the EMA weights.

\section{Post Hoc Polarity Ablation}
\label{app:polarity}
\textbf{Setup.} The ablation reuses the pre-specified conductivity generator, formula labels, hold-out split, and the pre-specified 400-label subsets. The inverted checkpoint is pretrained once and fine-tuned on subsets 0, 1, and 2 with seed 0 and on subset 0 with seeds 1--4, the cells on which the pre-specified models were evaluated. The only change in the matched-polarity run is that every synthetic pixel value $x$ is replaced by $255-x$ before preprocessing, so that the low-conductivity inclusions are bright as in the public data. Pretraining (12 epochs, learning rate $3\times10^{-4}$, batch 48, Smooth-L1) and fine-tuning (30 epochs, learning rate $5\times10^{-5}$, batch 32, head learning rate five times larger, dropout $0.1$, EMA $0.995$, mixup $0.3$, flips only, four-flip test-time averaging, last EMA weights) follow Section~\ref{sec:protocol} exactly. The augmentation run starts from the original mismatched checkpoint and applies $x\mapsto255-x$ to each training image with probability $0.5$ during fine-tuning only, on subset 0 with seed 0; the test images are not inverted.

\textbf{Polarity check.} On the 400 training images, the bright-phase area fraction correlates with the public $\kappa_{11}$ at $-0.970$; on the synthetic images, the dark-phase area fraction correlates with the formula label at $-0.991$, and after inversion the bright-phase fraction does so identically. The inversion therefore maps the formula's input to the same visual phase as the public target.

\begin{figure*}[t]
\begin{center}
\includegraphics[width=0.85\textwidth]{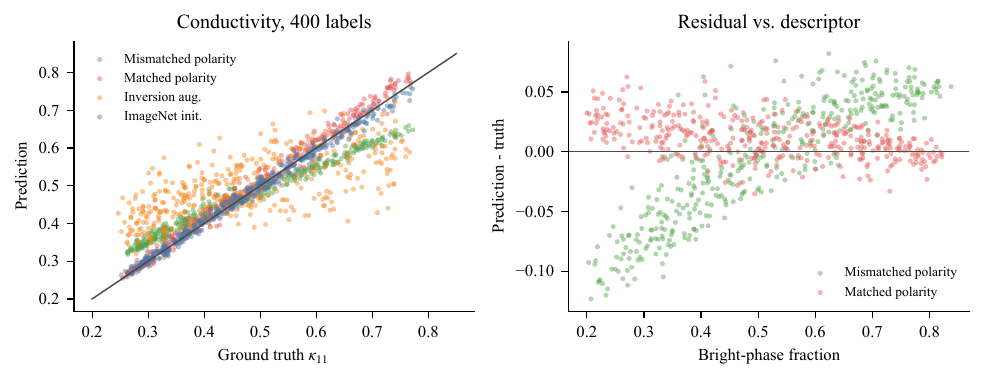}
\end{center}
\caption{Post hoc polarity ablation on conductivity (400 labels, subset 0, seed 0). Left: predictions against ground truth for the pre-specified formula model trained on mismatched-polarity synthetic images (green), the same pipeline with inverted synthetic images (red), the pre-specified checkpoint fine-tuned with 50\% random intensity inversion (orange), and ImageNet initialization (blue). Right: residual against the bright-phase area fraction for the two formula-pretrained models. Matching the polarity removes the compression and reverses the sign of the descriptor-dependent residual.}
\label{fig:polarity}
\end{figure*}

\textbf{Results.} Fig.~\ref{fig:polarity} shows the predictions, and Table~\ref{tab:polarity_full} gives the subset-0 metrics behind Table~\ref{tab:polarity} with the MAE and prediction ranges. The matched-polarity model covers the truth range ($[0.247,0.778]$) and slightly overshoots at the top ($0.796$), consistent with a slope of $1.03$; the mismatched and augmented models cover only $[0.309,0.662]$ and $[0.318,0.735]$. The sign change of the residual--descriptor correlation (from $+0.89$ to $-0.45$) indicates that the compression-type trend of the residual is replaced by a weaker trend of opposite sign.

\begin{table}[t]
\begin{center}
\footnotesize
\setlength{\tabcolsep}{3pt}
\resizebox{\linewidth}{!}{%
\begin{tabular}{|l|c|c|c|c|c|}
\hline
Model & RMSE & MAE & Slope & $r$ & Pred.\ range \\
\hline\hline
Mismatched & 0.0517 & 0.0442 & 0.64 & $+0.89$ & $[0.309,0.662]$ \\
Matched & 0.0190 & 0.0146 & 1.03 & $-0.45$ & $[0.247,0.796]$ \\
Mismatched + inv.\ aug. & 0.1104 & 0.0924 & 0.37 & $+0.85$ & $[0.318,0.735]$ \\
ImageNet & 0.0126 & 0.0096 & 0.96 & $+0.34$ & $[0.250,0.768]$ \\
\hline
\end{tabular}}
\end{center}
\caption{Full metrics of the polarity ablation (conductivity, 400 labels, subset 0, seed 0; truth range $[0.247,0.778]$). Mismatched and ImageNet are pre-specified models; Matched and the augmentation run are post hoc. $r$ is the correlation of the residual with the bright-phase area fraction.}
\label{tab:polarity_full}
\end{table}

\textbf{Replication.} Table~\ref{tab:polarity_rep} lists every fine-tuning of the inverted checkpoint. Across subsets the matched RMSE is $0.0170\pm0.0014$ (s.d.\ over subsets), against $0.0528\pm0.0009$ for the mismatched and $0.0133\pm0.0010$ for the ImageNet-initialized models on the same subsets. The 95\% paired-bootstrap interval of the RMSE difference, matched minus ImageNet, is $[+0.0057,+0.0072]$, $[+0.0007,+0.0021]$, and $[+0.0027,+0.0041]$ on subsets 0, 1, and 2, so the remaining gap to ImageNet is small on subset 1 but nonzero on all three; intensity histograms, inclusion shapes, and noise were not matched between the synthetic and public images. Seed 0 on subset 0, the main-text cell, ties with seed 2 as the worst of the five seeds. The five seeds have mean off-diagonal residual correlation $0.958$ and $N_{\mathrm{eff}}=1.07$; by Eq.~\eqref{eq:shared} their average retains $96.6\%$ of a single member's mean squared error, and the equal mean ($0.0172$) is worse than the best seed ($0.0151$).

\begin{table}[t]
\begin{center}
\footnotesize
\setlength{\tabcolsep}{4pt}
\begin{tabular}{|l|c|c|c|}
\hline
Run (matched polarity) & RMSE & Slope & $r$ \\
\hline\hline
Subset 0, seed 0 & 0.0190 & 1.03 & $-0.45$ \\
Subset 1, seed 0 & 0.0161 & 1.01 & $-0.30$ \\
Subset 2, seed 0 & 0.0160 & 1.02 & $-0.38$ \\
Mean $\pm$ s.d.\ over subsets & $0.0170\pm0.0014$ & & \\
\hline
Subset 0, seed 1 & 0.0151 & 1.03 & $-0.49$ \\
Subset 0, seed 2 & 0.0190 & 1.03 & $-0.47$ \\
Subset 0, seed 3 & 0.0174 & 1.03 & $-0.42$ \\
Subset 0, seed 4 & 0.0166 & 1.01 & $-0.31$ \\
Equal mean of seeds 0--4 & 0.0172 & & \\
\hline
\end{tabular}
\end{center}
\caption{Post hoc replication of the matched-polarity model (conductivity, 400 labels). One inverted pretraining run is fine-tuned on three label subsets and with five seeds on subset 0. The five seeds have $\bar\rho=0.958$ and $N_{\mathrm{eff}}=1.07$.}
\label{tab:polarity_rep}
\end{table}

\section{Ensemble Rules and Cross-Initialization Ensemble}
\label{app:crossinit}
Table~\ref{tab:rules} gives the pre-specified ensemble rules of Section~\ref{sec:collapse}. On conductivity the equal mean is worse than the best seed because averaging four weaker, nearly identical members with one stronger member cannot cancel a shared error. On stiffness the mean-alignment post-processing raises the error ($5.048$) because its $1.06\times$ inflation presupposes compressed predictions, whereas the stiffness slope is already $1.00$.

\begin{table}[t]
\begin{center}
\footnotesize
\setlength{\tabcolsep}{3pt}
\begin{tabular}{|l|c|c|}
\hline
Predictor (400 labels, subset 0) & Conductivity & Stiffness \\
\hline\hline
\multicolumn{3}{|l|}{Rules built on the five formula seeds} \\
\hline
Best formula seed & \textbf{0.0517} & 4.763 \\
Equal mean of 5 seeds & 0.0603 & 4.774 \\
$0.95\,\bar y+0.05\,y_{\mathrm{DINOv2}}$ & 0.0576 & \textbf{4.750} \\
Mixture + mean align, $1.06\times$ & 0.0526 & 5.048 \\
\hline
\multicolumn{3}{|l|}{Single models, no ensembling} \\
\hline
DINOv2 frozen + ridge alone & 0.0242 & 4.951 \\
Masked reconstruction (1 seed) & 0.0532 & 7.376 \\
ImageNet fine-tune (1 seed) & \textbf{0.0126} & \textbf{4.778} \\
\hline
$\bar\rho$ / $N_{\mathrm{eff}}$ of the 5 seeds & 0.994 / 1.01 & 0.999 / 1.00 \\
\hline
\end{tabular}
\end{center}
\caption{Test RMSE of ensemble rules built from five formula-pretrained seeds, compared with single models trained on the same 400 labels; bold marks the best entry of each block. The mixture weight, the alignment rule, and the ridge penalty were fixed before testing; all networks use the EMA weights.}
\label{tab:rules}
\end{table}

\textbf{Cross-initialization check.}
The check uses only the saved test predictions of the pre-specified 400-label, subset-0 models. Table~\ref{tab:crossinit} reports the participation ratio of the five formula seeds alone and with the ImageNet-initialized model added as a sixth member, and the equal-weight two-member means of formula seed 0 with the ImageNet and masked-reconstruction models. On stiffness the three initializations err on the same images (correlation $0.94$ with ImageNet), so adding a member changes $N_{\mathrm{eff}}$ by $0.03$; on conductivity the ImageNet model is decorrelated from the formula seed ($0.49$) but far more accurate, so the equal mean is dominated by the weaker member. Mixing seed 0 with the masked-reconstruction model, which is comparably accurate on conductivity ($0.0532$) and decorrelated ($0.29$), gives $0.0400$, the diversity gain available between comparably accurate members.

\textbf{Source of the small mixture gain.} With $\bar\rho\ge0.994$ the five seeds contribute no diversity in the sense of the bias--variance--diversity decomposition \cite{wood_2023_2}, so the $0.012$ gain of the mixture of Eq.~\eqref{eq:mix} over the best stiffness seed must be attributable to the DINOv2 member, whose residual correlates with the seed-mean residual at $0.875$ on stiffness and $0.263$ on conductivity (post hoc); the two-member form of Eq.~\eqref{eq:shared} reproduces the mixture RMSE exactly. The post hoc optimal weight, computed on test residuals and not used, is $0.37$ on stiffness (RMSE $4.680$) and $0.94$ on conductivity ($0.0239$): the only decorrelating member received almost no pre-specified weight, and on conductivity the best two-member ensemble essentially discards the synthetic-pretrained pipeline.

\begin{table}[t]
\begin{center}
\footnotesize
\setlength{\tabcolsep}{3pt}
\resizebox{\linewidth}{!}{%
\begin{tabular}{|l|c|c|}
\hline
 & Conductivity & Stiffness \\
\hline\hline
$\bar\rho$, five formula seeds & 0.994 & 0.999 \\
$N_{\mathrm{eff}}$, five formula seeds & 1.01 & 1.00 \\
$N_{\mathrm{eff}}$, five seeds + ImageNet & 1.28 & 1.03 \\
\hline
Formula seed 0 RMSE & 0.0517 & 4.795 \\
ImageNet RMSE & 0.0126 & 4.778 \\
Residual corr.\ (seed 0, ImageNet) & 0.49 & 0.94 \\
Equal mean (seed 0, ImageNet) & 0.0295 & \textbf{4.718} \\
Equal mean (seed mean, ImageNet) & 0.0336 & \textbf{4.711} \\
\hline
Masked-reconstruction RMSE & 0.0532 & 7.376 \\
Residual corr.\ (seed 0, masked) & 0.29 & 0.74 \\
Equal mean (seed 0, masked) & \textbf{0.0400} & 5.483 \\
\hline
\end{tabular}}
\end{center}
\caption{Post hoc cross-initialization ensemble on saved predictions (400 labels, subset 0). Bold marks the two-member means that improve on both of their members.}
\label{tab:crossinit}
\end{table}

\section{Shared Stiffness Outliers}
\label{app:outliers}
Table~\ref{tab:outliers} lists the ten test RVEs with the largest absolute residual of the formula-pretrained model at 1,600 labels (subset 0, seed 0), with the residuals of the other two initializations and the tabulated descriptors (fiber volume fraction, aspect ratio, fiber count). Nine of the ten are in the top ten of all three models, and the residual magnitudes agree to within a few GPa. The descriptor medians of the ten (volume fraction $28.8\%$, aspect ratio $2.68$, count $27$) are close to the test-set medians ($24.7\%$, $2.53$, $24$), and 27 of the 30 values lie inside the test-set 10th--90th percentile band ($12.9$--$36.9\%$; $1.31$--$3.71$; $17$--$33$); the exceptions are the volume fraction of mesh998 ($11.1\%$), the aspect ratio of mesh973 ($3.81$), and the fiber count of mesh981 ($34$). The released table contains no orientation descriptor, so the hypothesis that these RVEs have atypical fiber orientation distributions could not be tested from the tabulated data; measuring $\langle\cos^2\theta\rangle$ directly from the fiber ellipses in each image, as the synthetic generator does for its own fibers, would test it and is left for future work.

\begin{table}[t]
\begin{center}
\footnotesize
\setlength{\tabcolsep}{2.5pt}
\resizebox{\linewidth}{!}{%
\begin{tabular}{|l|r|r|r|r|r|r|}
\hline
RVE & Formula & ImageNet & Masked & $V_f$ (\%) & AR & $N_p$ \\
\hline\hline
mesh998 & 66.8 & 66.4 & 66.8 & 11.1 & 2.90 & 19 \\
mesh975 & 54.2 & 54.7 & 54.5 & 18.0 & 1.45 & 17 \\
mesh978 & 43.0 & 43.0 & 42.8 & 30.8 & 3.04 & 23 \\
mesh973 & 40.8 & 40.8 & 41.2 & 29.9 & 3.81 & 26 \\
mesh981 & 40.1 & 39.8 & 39.6 & 18.8 & 1.78 & 34 \\
mesh996 & 40.0 & 40.4 & 40.6 & 32.0 & 1.50 & 28 \\
mesh983 & 31.8 & 31.9 & 31.8 & 34.3 & 3.28 & 32 \\
mesh972 & 30.2 & 30.4 & 28.7 & 27.7 & 2.23 & 27 \\
mesh991 & 30.0 & 30.6 & 29.8 & 19.8 & 2.97 & 27 \\
mesh984$^{\dagger}$ & 22.7 & 22.6 & 19.9 & 32.7 & 2.46 & 30 \\
\hline
\end{tabular}}
\end{center}
\caption{Ten largest absolute residuals of the formula-pretrained stiffness model at 1,600 labels (columns 2--4: absolute residual in GPa for the formula, ImageNet, and masked-reconstruction models) with the tabulated descriptors ($V_f$: fiber volume fraction; AR: aspect ratio; $N_p$: fiber count). $^{\dagger}$Not in the top ten of the ImageNet or masked-reconstruction model.}
\label{tab:outliers}
\end{table}

\begin{table}[t]
\begin{center}
\footnotesize
\setlength{\tabcolsep}{3pt}
\resizebox{\linewidth}{!}{%
\begin{tabular}{|l|c|c|c|c|}
\hline
 & Subset 0 & Subset 1 & Subset 2 & Mean $\pm$ s.d. \\
\hline\hline
Conductivity RMSE & 0.0517 & 0.0519 & 0.0527 & $0.0521\pm0.0004$ \\
Conductivity $R^2$ & 0.864 & 0.862 & 0.858 & \\
Stiffness RMSE & 5.48 & 5.31 & 5.31 & $5.37\pm0.08$ \\
Stiffness $R^2$ & 0.939 & 0.943 & 0.942 & \\
\hline
\end{tabular}}
\end{center}
\caption{Frozen ImageNet ConvNeXt-Small features with ridge regression ($\alpha=50$) on the three 400-label subsets (post hoc). The DINOv2 ridge on the same subsets gives $0.0238\pm0.0002$ and $4.91\pm0.03$.}
\label{tab:imagenet_ridge}
\end{table}

\section{Frozen ImageNet Features}
\label{app:imagenet}
Table~\ref{tab:imagenet_ridge} gives the per-subset results of the frozen ImageNet ConvNeXt-Small ridge (Section~\ref{sec:frozen}). Features are the 768-dimensional global-average-pooled output of the timm backbone with ImageNet weights, computed with the same preprocessing as fine-tuning and without test-time augmentation; they are standardized on the 400 training images and fitted with ridge regression at the pre-specified penalty $\alpha=50$, as for the DINOv2 member.

\begin{table}[t]
\begin{center}
\footnotesize
\setlength{\tabcolsep}{3pt}
\resizebox{\linewidth}{!}{%
\begin{tabular}{|l|c|c|c|}
\hline
Labels & ImageNet & ViT-MAE & Formula \\
\hline\hline
\multicolumn{4}{|l|}{Conductivity $\kappa_{11}$ (test s.d.\ 0.140)} \\
\hline
100 & \textbf{0.0475} & 0.1125 & 0.2402 \\
400 (3 subsets) & $\mathbf{0.0190\pm0.0022}$ & $0.0341\pm0.0005$ & $0.0507\pm0.0023$ \\
1,600 & 0.0052 & \textbf{0.0041} & 0.0057 \\
\hline
\multicolumn{4}{|l|}{Stiffness $C_{1111}$ (test s.d.\ 22.1)} \\
\hline
100 & 16.27 & 16.14 & \textbf{5.38} \\
400 (3 subsets) & $4.87\pm0.09$ & $5.82\pm0.05$ & $\mathbf{4.70\pm0.02}$ \\
1,600 & 4.52 & \textbf{4.50} & 4.53 \\
\hline
\end{tabular}}
\end{center}
\caption{Post hoc ViT-S replication of Table~\ref{tab:curve}: test RMSE by label budget (fine-tuning seed 0; the 400-label row is the mean and s.d.\ over the same three subsets). The label-free control is a ViT masked autoencoder (ViT-MAE) started from ImageNet weights. All entries use the EMA weights. Bold marks the lowest RMSE in each row.}
\label{tab:vit}
\end{table}

\section{Second Backbone: ViT-S with a ViT Masked Autoencoder}
\label{app:vit}
To test whether the conclusions depend on the ConvNeXt backbone or on its weak control, the full budget grid was repeated post hoc with a ViT-S backbone (timm \texttt{vit\_small\_patch16\_224} with the \texttt{augreg\_in21k\_ft\_in1k} weights, position embeddings interpolated to $256\times256$ pixels, class-token readout, linear head) and with a ViT masked autoencoder \cite{he_2022_mae} as the label-free pretext: the encoder sees the 25\% visible patches, a four-block decoder (256 dimensions, 8 heads, fixed sine--cosine position embeddings) reconstructs the masked pixels under a mean-squared-error loss, and training starts from ImageNet weights with the formula-pretraining schedule. Every schedule, cell, seed, and metric was fixed before the run and is identical to the pre-specified protocol; no ViT-specific technique (e.g., layer-wise learning-rate decay) was added. The synthetic hold-out RMSE is $0.0039$ and $0.44$, against $0.0038$ and $0.44$ for ConvNeXt (run time in Appendix~\ref{app:env}).

Table~\ref{tab:vit} gives the budget curves. The ViT masked autoencoder is a much stronger control than dense ConvNeXt reconstruction: at 100 labels it reaches $0.113$ and $16.1$ where the ConvNeXt control reached $0.741$ and $24.6$, and at 1,600 labels it is the best initialization on both materials ($0.0041$ against $0.0052$ for ImageNet on conductivity, interval $[-0.0013,-0.0009]$; $4.50$ against $4.52$ on stiffness).

\textbf{Stiffness.} The formula initialization is best at 100 labels ($5.38$ against $16.27$ and $16.14$; intervals $[-12.1,-9.7]$ and $[-12.2,-9.4]$), remains better than ViT-MAE at 400 labels on every subset (upper interval bounds $-0.6$ to $-0.7$), is statistically tied with ImageNet at 400 labels ($4.70\pm0.02$ against $4.87\pm0.09$; the interval excludes zero on one subset of three), and all three reach $4.50$--$4.53$ at 1,600 labels. Five seeds from its checkpoint have $\bar\rho=0.999$, $N_{\mathrm{eff}}=1.00$, and RMSE $4.68$--$4.72$; their equal mean ($4.702$) is worse than the best seed ($4.678$).

\textbf{Conductivity.} The formula initialization is again less accurate than ImageNet at every budget ($0.0507\pm0.0023$ against $0.0190\pm0.0022$ at 400 labels; lower interval bound at least $+0.028$ on every subset) and shows the same compressed readout (slope $0.70$--$0.75$ at 400 labels; $-0.68$ at 100 labels, where the inverted polarity appears as an inverted prediction). Relative to the label-free pretext, however, the ordering is reversed: ViT-MAE outperforms the formula initialization at every budget ($0.113$ against $0.240$, $0.034$ against $0.051$, $0.0041$ against $0.0057$; lower bound at least $+0.011$ on every 400-label subset). Five seeds from its checkpoint do not collapse here: $\bar\rho=0.78$, $N_{\mathrm{eff}}=1.45$, RMSE $0.026$--$0.064$; their equal mean ($0.042$) is lower than the root mean square of the individual errors ($0.047$) but higher than the best seed ($0.026$), close to the $0.043$ of Eq.~\eqref{eq:shared} at $\bar\rho=0.78$, $M=5$.

\textbf{Interpretation.} Under rules fixed before the run and with the EMA, the replication (i) reproduces the polarity confound and the conductivity ordering against ImageNet, (ii) reproduces the near-unit stiffness seed correlation with a second backbone, which Section~\ref{sec:collapse} attributes to the task, and (iii) makes the conductivity ordering of formula label and label-free pretext depend on both polarity and the EMA: with aligned images the formula label leads with the EMA (Appendix~\ref{app:followup}) but trails ViT-MAE without it (Appendix~\ref{app:ema}).

\begin{table}[t]
\begin{center}
\footnotesize
\setlength{\tabcolsep}{3pt}
\resizebox{\linewidth}{!}{%
\begin{tabular}{|l|c|c|c|c|}
\hline
 & Subset 0 & Subset 1 & Subset 2 & Mean $\pm$ s.d. \\
\hline\hline
\multicolumn{5}{|l|}{Matched-polarity RMSE by pretraining run (fine-tuning seed 0)} \\
\hline
Pretraining run 0 & 0.0190 & 0.0161 & 0.0160 & $0.0170\pm0.0014$ \\
Pretraining run 1 & 0.0208 & 0.0187 & 0.0189 & $0.0195\pm0.0009$ \\
Pretraining run 2 & 0.0206 & 0.0210 & 0.0208 & $0.0208\pm0.0002$ \\
\hline
\multicolumn{5}{|l|}{Across the three checkpoints (one model per checkpoint)} \\
\hline
$\bar\rho$ between checkpoints & 0.895 & 0.879 & 0.882 & \\
$N_{\mathrm{eff}}$ & 1.15 & 1.17 & 1.17 & \\
Equal mean RMSE & 0.0194 & 0.0180 & 0.0179 & \\
Best checkpoint RMSE & 0.0190 & 0.0161 & 0.0160 & \\
\hline
\end{tabular}}
\end{center}
\caption{Post hoc pretraining-seed replication of the matched-polarity ConvNeXt model (conductivity, 400 labels) and the residual dependence between models fine-tuned from different pretraining runs.}
\label{tab:pretrain_seeds}
\end{table}

\begin{table}[t]
\begin{center}
\footnotesize
\setlength{\tabcolsep}{3pt}
\resizebox{\linewidth}{!}{%
\begin{tabular}{|l|c|c|}
\hline
 & Conductivity & Stiffness \\
\hline\hline
Mixed polarity, 100 labels & 0.0353 & 5.540 \\
Reference, 100 labels & 0.0351 & 5.492 \\
ImageNet, 100 labels & 0.0723 & 13.28 \\
\hline
Mixed polarity, 400 labels (3 subsets) & $0.0192\pm0.0004$ & $4.72\pm0.06$ \\
Reference, 400 labels (3 subsets) & $0.0170\pm0.0014$ & $4.78\pm0.07$ \\
ImageNet, 400 labels (3 subsets) & $0.0133\pm0.0010$ & $4.84\pm0.05$ \\
\hline
Mixed polarity, 1,600 labels & 0.0057 & 4.487 \\
Reference, 1,600 labels & 0.0053 & 4.487 \\
ImageNet, 1,600 labels & 0.0041 & 4.492 \\
\hline
Mixed: slope at 400 labels & 0.99--1.00 & 0.99--1.00 \\
Mixed: $\bar\rho$ / $N_{\mathrm{eff}}$ of five seeds & 0.997 / 1.00 & 0.999 / 1.00 \\
Mixed: equal mean / best seed & 0.0190 / 0.0188 & 4.732 / 4.698 \\
Mixed: synthetic hold-out RMSE & 0.0039 & 0.445 \\
\hline
\end{tabular}}
\end{center}
\caption{Post hoc mixed-polarity pretraining (ConvNeXt-Small; each synthetic training image inverted with probability 0.5). Reference is the matched-polarity model for conductivity and the pre-specified formula model for stiffness. The synthetic hold-out is evaluated at the original polarity. All entries use the EMA weights.}
\label{tab:mixed}
\end{table}

\section{Post Hoc Follow-Up Runs}
\label{app:followup}
\textbf{Rules.} Four runs were added after the polarity ablation. For each, the configuration, the cells, and a decision rule were fixed before it was run, and no setting was changed after its test errors were seen. All use the pre-specified schedules, the three 400-label subsets, five fine-tuning seeds on subset 0, and paired bootstrap intervals with 4,000 resamples; the pre-specified predictions and checkpoints are only read.

\textbf{Pretraining runs.} Rule: if the three-subset means of two further inverted pretrainings are clearly below the mismatched model and above ImageNet, neither the correction nor the remaining gap depends on one pretraining run. With the EMA both runs satisfy it (Table~\ref{tab:pretrain_seeds}; without the EMA they do not, Appendix~\ref{app:ema}): on every subset the matched-minus-ImageNet intervals lie within $[+0.0031,+0.0092]$ and the matched-minus-mismatched intervals within $[-0.0366,-0.0294]$. Both end at a synthetic hold-out RMSE of $0.0038$, and their five seeds on subset 0 have $\bar\rho=0.984$ and $0.987$, $N_{\mathrm{eff}}=1.03$ and $1.02$.

\textbf{Cross-checkpoint dependence.} Rule, fixed before the correlations were computed from saved predictions: between-checkpoint $\bar\rho\le0.90$ on all three subsets and clearly below the within-checkpoint value means that the collapse arises mainly from sharing a checkpoint; $\bar\rho\ge0.95$ with $N_{\mathrm{eff}}\le1.2$ means that it does not. The first case holds (Table~\ref{tab:pretrain_seeds}). On subset 0, the 15 models from three checkpoints and five fine-tuning seeds have mean correlation $0.976$ within and $0.871$ between checkpoints (minimum $0.716$) and $N_{\mathrm{eff}}=1.20$; their equal mean ($0.0181$) is worse than the best member ($0.0151$). These correlations use the EMA weights; without the EMA the within-checkpoint value of the first checkpoint falls from $0.958$ to $0.873$ (Appendix~\ref{app:ema}), and the between-checkpoint values were not recomputed.

\textbf{Polarity-corrected ViT-S.} Both the formula and the ViT masked-autoencoder pretraining use the inverted synthetic images; everything else follows Appendix~\ref{app:vit}. Rule: the primary comparison is the inverted formula minus the inverted ViT-MAE at 400 labels, and upper bounds below zero on all three subsets mean that the label outperforms the label-free pretext on aligned images. The intervals are $[-0.0154,-0.0132]$, $[-0.0112,-0.0089]$, and $[-0.0131,-0.0108]$, so the rule is met with the EMA; without it the order reverses (Appendix~\ref{app:ema}). Against ImageNet the inverted formula is worse on subsets 0 and 1 ($[+0.0034,+0.0050]$, $[+0.0059,+0.0078]$), better on subset 2 ($[-0.0024,-0.0008]$), better at 100 labels ($[-0.0138,-0.0111]$), and marginally better at 1,600 labels ($[-0.0004,-0.00001]$), where the inverted ViT-MAE is best ($0.0039$). Inverting the ViT-MAE images changes its 400-label error by at most $0.002$, with signs that differ across subsets. The synthetic hold-out RMSE ends at $0.0039$. Its five seeds have $\bar\rho=0.984$ and $N_{\mathrm{eff}}=1.03$, with equal mean $0.0221$ against best seed $0.0215$, whereas the mismatched ViT conductivity checkpoint had $N_{\mathrm{eff}}=1.45$.

\textbf{Mixed polarity.} Rules: on conductivity at 400 labels, mixed minus mismatched with upper bounds below zero on all three subsets means that mixed pretraining mitigates the polarity error; mixed minus matched including zero on at least two subsets means ``similar to matched'', and lower bounds above zero on all three means ``worse than matched''. On stiffness, mixed minus the pre-specified formula including zero or favoring mixed on at least two of the three 400-label subsets and at 100 labels means ``no harm''. Mixed pretraining is deemed a practical method that needs no polarity check only if mitigation and no harm both hold. Table~\ref{tab:mixed} gives the errors. Mitigation holds (upper bounds $-0.031$ to $-0.034$). Against the matched model, subset 0 includes zero ($[-0.0005,+0.0005]$) while subsets 1 and 2 do not ($[+0.0022,+0.0032]$, $[+0.0033,+0.0042]$), so neither label applies. On stiffness all three 400-label subsets favor mixed (upper bounds $-0.045$, $-0.018$, $-0.042$), but at 100 labels mixed is worse ($[+0.039,+0.061]$), so the no-harm criterion fails and the combined criterion is not met; at 1,600 labels the interval includes zero. Inclusion fractions of the conductivity generator span $0.18$--$0.84$, so brightness alone does not identify the inclusion phase and the mixed-polarity network must use other cues; its error of $0.0192$ is consistent with shape cues, which were not tested directly.

\section{EMA-Free Recheck}
\label{app:ema}
\textbf{Rules.} The recheck was planned after all other results were seen and run in five stages, each fixed before its own results were seen: 100 labels, then 400 labels, then five fine-tuning seeds of three formula checkpoints at 400 labels, then five ImageNet seeds on stiffness, and finally the extension described under \emph{Final stage}. The only change is that no EMA shadow is kept and the last weights of epoch 30 are evaluated with the same four-flip test-time averaging; schedules, learning rates, heads, augmentation, splits, and checkpoints are those of Section~\ref{sec:protocol}, and the polarity-matched cells use the inverted checkpoint of Section~\ref{sec:polarity}. Three comparisons were judged, each as formula minus ImageNet: stiffness, mismatched conductivity, and matched conductivity. If the sign of the difference without the EMA is opposite to the sign with the EMA, the statement is reported as EMA-dependent; if the signs agree and the 95\% paired-bootstrap interval (4,000 resamples, seed 0) of the EMA-free difference includes zero, as a tie; otherwise as the same conclusion. For the formula seeds, $\bar\rho\ge0.95$ without the EMA means that the collapse holds, $\bar\rho\le0.90$ that it depends largely on the EMA, and values in between are reported as intermediate. For the stiffness ImageNet seeds, $\bar\rho\ge0.95$ means that the collapse is an error structure of the task, $\bar\rho\le0.90$ that it is specific to the formula checkpoint, and values in between are intermediate.

\textbf{Outcome.} At 100 labels, stiffness ($[-0.98,-0.38]$) and mismatched conductivity ($[+0.017,+0.020]$) give the same conclusion as with the EMA, and matched conductivity is EMA-dependent ($[+0.0044,+0.0064]$, against $0.0351$ versus $0.0723$ with the EMA). At 400 labels, mismatched conductivity gives the same conclusion ($[+0.0024,+0.0033]$) and matched conductivity is EMA-dependent ($[-0.0009,-0.0001]$); the stiffness difference changes sign from $+0.017$ to $-0.019$, which the first rule labels EMA-dependent, although both intervals include zero ($[-0.06,+0.02]$ without the EMA). The five seeds of Table~\ref{tab:ema_seeds} resolve it: their ranges separate in favor of the formula ($4.491$--$4.516$ versus $4.535$--$4.638$), by the seed-range criterion of the final stage. We note that this resolution is post hoc in two senses: the seed-range criterion was specified for the conductivity final-stage block~(A), not for stiffness, and it was applied here only after the stiffness seed ranges of Table~\ref{tab:ema_seeds} had already been seen. It also conflicts with the first rule above, which from the sign flip alone would call the stiffness comparison EMA-dependent; we report both verdicts and let the reader weigh them. Without the EMA the slopes of prediction on truth are $0.91$--$0.98$ at 100 labels and $0.97$--$1.01$ at 400 labels for all seven cells, and the residual correlation with the bright-phase fraction at 400 labels is $+0.32$ (mismatched), $+0.11$ (matched), $+0.16$ (ImageNet), and $-0.18$ (masked reconstruction). Table~\ref{tab:ema_seeds} gives the seed ensembles. The stiffness formula collapse holds, but the ImageNet seeds reach $\bar\rho=0.992$, so by the last rule it is an error structure of the task; formula and ImageNet seeds correlate at $0.974$ on average (minimum $0.963$), and without the ten largest formula residuals the two within-initialization values are $0.989$ and $0.952$. Both conductivity checkpoints fall in the EMA-dependent range, the matched one ($0.873$) close to the boundary of $0.90$.

\begin{table}[t]
\begin{center}
\footnotesize
\setlength{\tabcolsep}{3pt}
\resizebox{\linewidth}{!}{%
\begin{tabular}{|l|c|c|c|c|}
\hline
 & Stiff., formula & Stiff., ImageNet & Cond., mismatched & Cond., matched \\
\hline\hline
$\bar\rho$ with EMA & 0.999 & -- & 0.994 & 0.958 \\
$\bar\rho$ without EMA & 0.998 & 0.992 & 0.734 & 0.873 \\
Smallest $\rho_{ij}$ without EMA & 0.997 & 0.987 & 0.497 & 0.822 \\
$N_{\mathrm{eff}}$ without EMA & 1.00 & 1.01 & 1.55 & 1.23 \\
Seed RMSE range & 4.491--4.516 & 4.535--4.638 & 0.0113--0.0130 & 0.0077--0.0088 \\
Equal mean RMSE & 4.500 & 4.566 & \textbf{0.0108} & 0.0079 \\
Best seed RMSE & \textbf{4.491} & \textbf{4.535} & 0.0113 & \textbf{0.0077} \\
\hline
\end{tabular}}
\end{center}
\caption{Five fine-tuning seeds without the EMA (400 labels, subset 0; post hoc). Bold marks the better of the equal mean and the best seed; --: not run.}
\label{tab:ema_seeds}
\end{table}

\textbf{Final stage.} Three blocks, without the EMA, with seed 0 reused from the earlier stages where it exists. (A) Conductivity at 400 labels on all three subsets with five seeds, for the matched pretraining runs, the mismatched pretraining runs, and ImageNet. The mismatched side had a single run at this stage; it was completed to three runs by a later pre-specified replication (Appendix~\ref{app:mismatched}), whose run-summary rule the verdict below follows. A run summary is the mean over the five fine-tuning seeds of a subset, and a run's overall summary is the mean of its three subset summaries; ranges are separated when the maximum of one is below the minimum of the other. The polarity effect \emph{holds} only if the matched run summaries separate below the mismatched run summaries both overall and on all three subsets, and is \emph{not separated} if they overlap or reverse overall or on two or more subsets. (B) Stiffness at 100 labels on subset 0 with five seeds, the formula checkpoint with its head kept or re-initialized (encoder loaded only) and ImageNet. If both formula arms lie below ImageNet, the encoder carries the advantage; if only the kept arm does, the head does. (C) ViT-S conductivity at 400 labels, seed 0, matched formula (pretrained head kept) against matched ViT-MAE and against ImageNet (new heads); bootstrap upper bounds below zero on all three subsets mean that the formula label is better, lower bounds above zero that it is worse.

Table~\ref{tab:ema_final} (moved to the main text as the primary EMA-free comparison) gives the outcome (3,477\,s, 228\,s, and 335\,s). (A) The matched run summaries ($0.0086$/$0.0111$/$0.0118$) separate just below the mismatched summaries ($0.0119$/$0.0120$/$0.0123$) overall, but their per-subset ranges overlap on subsets 0 and 2 and separate only on subset 1, so by the combined run-summary rule the verdict is \emph{not separated} (Appendix~\ref{app:mismatched}); the overall separation is the most extreme outcome attainable at three runs per side (one-sided permutation $p=1/20$), so this is inconclusive rather than evidence of no effect. (B) Both formula arms lie below ImageNet, so the encoder carries the advantage. (C) The formula label is worse than ViT-MAE ($[+0.0030,+0.0037]$, $[+0.0011,+0.0017]$, $[+0.0011,+0.0018]$) despite its pretrained head, and better than ViT ImageNet ($[-0.0061,-0.0052]$, $[-0.0022,-0.0014]$, $[-0.0070,-0.0059]$).

\section{Mismatched-Polarity Pretraining Replication}
\label{app:mismatched}
At the final stage of the EMA-free recheck the matched (inverted) polarity had three pretraining runs but the mismatched (original) polarity had only the pre-specified checkpoint, so a single mismatched run was being compared with the spread of three matched runs. This replication completes the mismatched side so that the contrast-polarity effect is tested against the same three-run variation on both sides, at the EMA-free 400-label conductivity cells.

\textbf{Setup.} Mismatched run 0 is the pre-specified non-inverted checkpoint (seed 0, no inversion, no $90^\circ$ rotation), reused without retraining, and its fine-tunings are read from the EMA-free stages above. Mismatched runs 1 and 2 are new pretrainings of the same synthetic images and formula labels with pretraining seeds 1 and 2, not inverted, and otherwise the pre-specified schedule (12 epochs, learning rate $3\times10^{-4}$, batch 48). Each new checkpoint keeps its pretrained head and is fine-tuned without the EMA on the three 400-label subsets with five fine-tuning seeds, using the last weights of epoch 30 and the four-flip test-time averaging of Section~\ref{sec:protocol}. The configuration and the decision rule were fixed before the runs in an internal protocol file; only the missing cells were trained, and no pre-existing file was modified. Pretraining seed 1 took 6,000\,s and its fifteen fine-tunings 818\,s; pretraining seed 2 was interrupted at epoch 11 and resumed (the resumed segment 558\,s, the full run comparable to seed 1), and its fine-tunings took 816\,s, so the two new checkpoints and their fine-tunings cost on the order of 13,600\,s in total, comparable to the inverted two-pretraining follow-up (Appendix~\ref{app:env}). The driver script's own wall time was 1,375\,s, but that figure reflects only its final invocation, which read the cached seed-1 checkpoint and resumed seed-2 from epoch 11 and therefore excludes most of the pretraining compute.

\textbf{Rule.} The decision uses run summaries only, where a subset summary is the mean of the five fine-tuning seeds and a run's overall summary is the mean of its three subset summaries. Two sets of summaries are separated when the maximum of one is below the minimum of the other. The effect is \emph{polarity holds} only if the matched summaries separate below the mismatched summaries both overall and on all three subsets, \emph{reversed} if the mismatched summaries separate below overall, and \emph{not separated} otherwise; the final verdict is \emph{not separated} whenever either the overall or the per-subset comparison is not separated. Fine-tuning-seed ranges, paired bootstraps, the ImageNet comparison, and the synthetic hold-out are recorded but excluded from the verdict.

\textbf{Result.} The run summaries are in Table~\ref{tab:ema_final}, block~(A): matched $0.0086$/$0.0111$/$0.0118$ and mismatched $0.0119$/$0.0120$/$0.0123$. Overall the matched maximum ($0.0118$) is just below the mismatched minimum ($0.0119$), so the overall comparison alone would read \emph{polarity holds}; per subset, however, the matched and mismatched summaries overlap on subsets 0 and 2 and separate (matched below) only on subset 1. The final verdict is therefore \emph{not separated}. This is inconclusive rather than evidence of no effect: with three runs per side the one-sided permutation test on the run summaries has a smallest attainable $p$-value of $1/20=0.05$, so even the most extreme outcome is only weak evidence, and here the overall comparison already reaches that extreme (the matched mean $0.0105$ lies below the mismatched mean $0.0121$; the matched maximum $0.011794$ is below the mismatched minimum $0.011936$ before rounding). The effect is also heterogeneous: the matched runs span $0.0086$--$0.0118$ while the mismatched runs cluster tightly ($0.0119$--$0.0123$), so matching the polarity helps on some pretraining runs and not on others. The two new mismatched checkpoints end at synthetic hold-out RMSE $0.00385$ and $0.00388$, matching the other pretrainings, so their fine-tuning errors do not stem from a worse pretext fit. Against ImageNet (recorded, not part of the verdict), by fine-tuning-seed range the mismatched runs 1 and 2 are worse on all three subsets and run 0 is mixed across subsets; no mismatched run beats ImageNet.

\onecolumn
	\begin{center}
		\section*{Summary of this paper}
		\setcounter{subsection}{0}
	\end{center}
\noindent\fbox{
    \parbox{\textwidth}{
    	\begin{quote}
		\subsection{Problem Setup}
		Image-to-scalar regression from micrographs to a material property is limited by the number of labeled images, each requiring a destructive test or a simulation. Closed-form homogenization formulas label synthetic microstructures exactly and at no cost, making supervised pretraining on formula labels an alternative to self-supervised pretraining. We ask two questions with one pre-specified protocol on two public benchmarks (two-phase conductivity $\kappa_{11}$, 2,000 fixed test images; short-fiber stiffness $C_{1111}$, the official 1,000-image validation set): does a formula label transfer better than a label-free pretext or ImageNet initialization at 100, 400, or 1,600 real labels, and how does an ensemble behave when all members learned the same formula?
		\vspace{0.2\baselineskip}
		\subsection{Novelty}
		(i) A controlled two-material comparison of formula-labeled pretraining against masked reconstruction on the same synthetic images and against ImageNet initialization, with every hyperparameter fixed before testing. (ii) Seed dependence under a shared formula checkpoint: with the pre-specified exponential moving average (EMA), five fine-tuning seeds have $\bar\rho$ of 0.994 and 0.999; post hoc without the EMA, $\bar\rho$ falls to 0.73 on conductivity and stays at 0.998 on stiffness, where ImageNet seeds also reach 0.992, so the stiffness value reflects the task. (iii) Two post hoc confounds of the conductivity failure: the generator inverts the contrast polarity of the benchmark, and the EMA (decay 0.995, no bias correction) still holds 64\% and 16\% of the initial weights at 100 and 400 labels. The compressed readout (slope 0.64) exists only in the EMA weights. Without the EMA the mismatched checkpoint has $1.4$ times the ImageNet error (0.0119 versus 0.0084), and with three pretraining runs per polarity the matched means (0.0086, 0.0111, 0.0118) only narrowly separate from the mismatched (0.0119--0.0123) and overlap on two of three subsets, so a pre-specified rule finds the effect inconclusive and heterogeneous. (iv) Low-cost baselines, an affine-calibrated formula and ridge regression on frozen DINOv2 features, that beat the EMA-averaged mismatched-polarity networks but not the EMA-free ones.
		\vspace{0.2\baselineskip}
		\subsection{Algorithms}
		Per material, 150,000 synthetic images are labeled by a two-dimensional Maxwell Garnett expression or an orientation-averaged Halpin--Tsai surrogate. A ConvNeXt-Small backbone is pretrained for 12 epochs at learning rate $3\times10^{-4}$ either by regressing the formula label (Smooth-L1) or by masked reconstruction; each initialization is fine-tuned for 30 epochs at $5\times10^{-5}$ (AdamW, mixup, weight EMA, four-flip test-time averaging), and the 100- and 400-label cells are refitted post hoc with the EMA disabled. Ensemble rules on five formula seeds: equal mean, the mixture $\hat y=0.95\,\bar y+0.05\,y_{\mathrm{DINOv2}}$ (ridge $\alpha=50$ on frozen DINOv2 features), and mean alignment. Dependence is the participation ratio $N_{\mathrm{eff}}=(\sum\lambda_j)^2/\sum\lambda_j^2$ of the residual correlation matrix. Formula baselines evaluate the formula on real-image descriptors, zero-shot and with an affine fit.
		\vspace{0.2\baselineskip}
		\subsection{Experiments}
		Stiffness: formula pretraining gives RMSE 5.49 at 100 labels versus 13.28 (ImageNet) and 24.63 (masked reconstruction), ties the three at 400 ($4.78\pm0.07$ versus $4.84\pm0.05$) and 1,600 labels ($\approx$4.49). Conductivity: ImageNet is best at every pre-specified budget (0.0723 / $0.0133\pm0.0010$ / 0.0041), formula pretraining gives 0.188 / $0.0528\pm0.0009$ / 0.0055 and masked reconstruction 0.741 / $0.0503\pm0.0027$ / 0.0064. Baselines: affine-calibrated formula 0.0334 and 4.84; frozen DINOv2 ridge $0.0238\pm0.0002$ and $4.91\pm0.03$. Without the EMA the small-budget errors shrink sharply (conductivity masked $0.741\!\to\!0.054$ at 100 labels), and the stiffness formula beats ImageNet over five seeds at 100 labels (4.58 versus 5.55; 4.60 with a re-initialized head). Matched polarity gives $0.0170$--$0.0208$ at 400 conductivity labels with the EMA; without it the matched (0.0086--0.0118) and mismatched (0.0119--0.0123) runs are not cleanly separated (inconclusive, heterogeneous). In a post hoc ViT-S run, the matched formula label beats a ViT masked autoencoder with the EMA but not without it (one seed).
		\end{quote}
    }
}
\thispagestyle{empty}
\end{document}